%% file: main.tex
\documentclass[10pt,twocolumn,letterpaper]{article}

\usepackage{cvpr}              

\usepackage{amsmath}
\usepackage{amssymb}
\usepackage{gensymb}
\usepackage{nicefrac}
\usepackage{fontawesome}
\usepackage{pifont}
\usepackage{textcomp}
\newcommand{\cmark}{\ding{51}}%
\newcommand{\xmark}{\ding{55}}%

\usepackage{booktabs}
\usepackage{colortbl}
\usepackage{array}
\usepackage{ragged2e}
\usepackage{tabularx}

\usepackage{makecell}
\usepackage{multirow}
\usepackage{xspace}

\usepackage{graphicx}
\usepackage{siunitx}
\newcommand{\qty}[2]{\SI{#1}{#2}}
\newcommand{\numproduct}[1]{\num{#1}}
\newcommand{\unit}[1]{\si{#1}}
\usepackage[table,dvipsnames]{xcolor}
\usepackage{microtype}
\usepackage[accsupp]{axessibility}
\makeatletter
\newcommand{\settitle}{\@maketitle}
\makeatother

\usepackage[numbers,sort,compress]{natbib}
\makeatletter
\def\NAT@spacechar{~}
\makeatother

\usepackage{enumitem}
\setlist{nosep}
\renewcommand{\paragraph}[1]{\par\vspace{2pt plus 1pt minus 1pt}\noindent{\bfseries #1\enspace}}
\usepackage[pagebackref,breaklinks,colorlinks]{hyperref}

\usepackage[capitalize]{cleveref}
\crefname{equation}{Eq.}{Eqs.}
\Crefname{equation}{Eq.}{Eqs.}
\crefname{section}{Sec.}{Secs.}
\Crefname{section}{Sec.}{Secs.}
\Crefname{table}{Tab.}{Tabs.}
\crefname{table}{Tab.}{Tabs.}

\def\cvprPaperID{2145} 
\def\confName{CVPR}
\def\confYear{2023}

\newcommand{\greencheck}{{\color{Green}\cmark}\xspace}
\newcommand{\green}{\cellcolor{Green!12.5}\greencheck}
\newcommand{\yellowcheck}{{\color{YellowOrange}(\cmark)}\xspace}
\newcommand{\yellow}{\cellcolor{YellowOrange!12.5}\yellowcheck}
\newcommand{\redcheck}{{\color{red}\xmark}\xspace}
\newcommand{\red}{\cellcolor{red!12.5}\redcheck}
\newcommand{\lock}{\faLock\xspace}
\newcommand{\unlock}{\textcolor{NavyBlue}{\faUnlock}\xspace}
\newcommand{\redstar}{{\color{red}{*}}\xspace}
\renewcommand{\eg}{e.\@g.\@}
\renewcommand{\ie}{i.\@e.\@}

\begin{document}

\title{The Differentiable Lens: \\ Compound Lens Search over Glass Surfaces and Materials for Object Detection}

\author{
    Geoffroi Côté$^{1,2}$
    \quad Fahim Mannan$^3$
    \quad Simon Thibault$^1$
    \quad Jean-François Lalonde$^1$
    \quad Felix Heide$^{2,3}$
    \\
    $^1$Université Laval
    \quad $^2$Princeton University
    \quad $^3$Algolux\\
}
\twocolumn[\settitle]

\begin{abstract}
    Most camera lens systems are designed in isolation, separately from downstream computer vision methods.
    Recently, joint optimization approaches that design lenses alongside other components of the image acquisition and processing pipeline---notably, downstream neural networks---have achieved improved imaging quality or better performance on vision tasks.
    However, these existing methods optimize only a subset of lens parameters and cannot optimize glass materials given their categorical nature.
    In this work, we develop a differentiable spherical lens simulation model that accurately captures geometrical aberrations.
    We propose an optimization strategy to address the challenges of lens design---notorious for non-convex loss function landscapes and many manufacturing constraints---that are exacerbated in joint optimization tasks.
    Specifically, we introduce quantized continuous glass variables to facilitate the optimization and selection of glass materials in an end-to-end design context, and couple this with carefully designed constraints to support manufacturability.
    In automotive object detection, we report improved detection performance over existing designs even when simplifying designs to two- or three-element lenses, despite significantly degrading the image quality.
\end{abstract}

\input{paper.tex}

{\small\bibliography{biblio}}

\clearpage

\title{Supplementary Information -- The Differentiable Lens: \\ Compound Lens Search over Glass Surfaces and Materials for Object Detection}

\maketitle
\thispagestyle{empty}

\renewcommand{\theequation}{S\arabic{equation}}
\renewcommand{\thefigure}{S\arabic{figure}}
\renewcommand{\thetable}{S\arabic{table}}
\renewcommand{\thesection}{S\arabic{section}}
\setcounter{equation}{0}
\setcounter{figure}{0}
\setcounter{table}{0}
\setcounter{section}{0}
\input{supplement.tex}

\end{document}

%% file: paper.tex
\begin{figure}[t]
    \centering
    \includegraphics{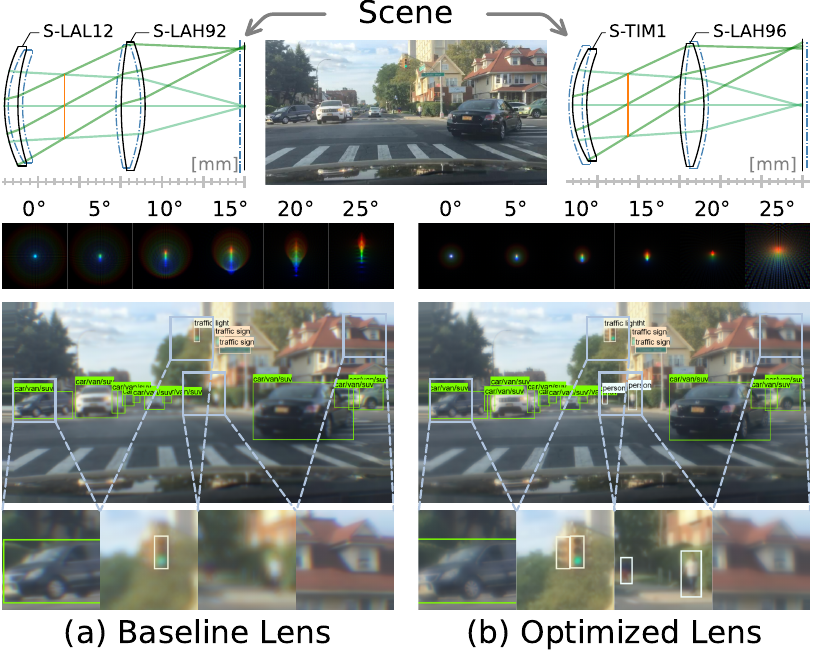}
    \caption{
        We introduce a differentiable lens simulation model and an optimization method to optimize compound lenses specifically for downstream computer vision tasks, and apply them to automotive object detection.
        Here, although the optimized two-element lens has a worse average spot size than the baseline lens (\qty{136}{\um} vs \qty{80}{\um}), it achieves a better mean average precision (AP) on the BDD100K dataset (32.0 vs 30.3).
        The optimized lens sacrifices optical performance near the corners for better performance in the small and medium field values where most of the objects are located.
        In lens layout plots, dashed lines represent the baseline/optimized counterpart and annotations indicate the optimized glass materials.
        }
    \label{fig:teaser}
\end{figure}

\section{Introduction}

The prevailing design paradigm for typical optical systems is to conceive them in isolation by use of simplified image quality metrics such as spot size~\cite{smith2004modern}.
However, achieving ideal imaging properties or optimal performance on computer vision tasks generally requires a more comprehensive approach that includes the remaining parts of the image acquisition and processing chain, in particular the sensor, image signal processing, and downstream neural networks.

Over the years, many works have addressed the joint design of simple optical systems such as diffractive optical elements (DOEs)~\cite{chang2019deep,metzler2020deep,sitzmann2018end,sun2020learning}.
These works approach joint optics design by simplifying the design to a single phase plate that allows for a differentiable paraxial Fourier image formation model, optimizable via stochastic gradient descent (SGD) variants.
More recently, several differentiable lens simulation models have been introduced to address the more complex compound lens systems present in most commodity-type cameras.
\citet{tseng2021differentiable} build such a model by training a proxy neural network, whereas other works~\cite{sun2021end,hale2021end,li2021end} directly implement differentiable ray-tracing operations in automatic differentiation frameworks~\cite{paszke2019pytorch,i2016tensorflow}, an idea also discussed in~\cite{volatier2017generalization,wang2021lens,cote2021on}.
However, all relevant previous works~\cite{sun2021end,hale2021end,li2021end,tseng2021differentiable} optimize over only a subset of possible surface profiles and spacings, and ignore the optimization of glass materials altogether.
Yet, allowing \emph{all} lens variables to be freely optimized---that is, without predefined boundaries---provides an opportunity for increased performance on downstream tasks.

Unfortunately, lens design optimization is no trivial process.
Even optimizing for traditional optical performance metrics presents significant difficulties, notably: harsh loss function landscapes with abundant local minima and saddle points~\cite{sturlesi1991future,vanturnhout2009chaotic,turnhout2009instabilities}, restrictive manufacturing constraints~\cite{smith2004modern,bentley2012field}, and risk of ray-tracing failures.
Optimizing a lens jointly on vision tasks only exacerbates these pitfalls due to the noisy gradients of SGD when applied to complex vision models~\cite{tseng2021differentiable}.
Moreover, joint optimization does not naturally allow external supervision from lens designers and, as such, does not necessarily result in a manufacturable lens.

In this work, we introduce a computationally efficient and differentiable pipeline for simulating and differentiating through compound spherical refractive lenses with respect to all design parameters in an end-to-end manner.
Our forward model integrates exact optical ray tracing, accurate ray aiming, relative illumination, and distortion.
Furthermore, we develop an optimization strategy to facilitate the end-to-end design of refractive lenses using SGD-based optimizers while strongly encouraging manufacturable outcomes.
To this end, we carefully define losses to handle design constraints, and introduce \emph{quantized continuous glass variables} to facilitate the process of selecting the best glass materials among glass catalogs that contain dozens of candidates---a challenge unmet in prior joint optimization methods.

We apply our simulation and optimization pipeline to the task of object detection (OD).
We find that even simple two-element lenses such as the ones in \cref{fig:teaser} can be compelling candidates for low-cost automotive OD despite a noticeably worse image quality.
Then, we validate the proposed method by demonstrating that optimizing the lens jointly with the OD model leads to consistent improvements in detection performance.
We make the following contributions:

\begin{itemize}
    \item We introduce a novel method for simulating and optimizing compound optics with respect to glass materials, surface profiles, and spacings.
    \item We validate the method on the end-to-end optimization of an OD downstream loss, with lenses specifically optimized for intersection over union (IoU) of bounding boxes predicted from a jointly trained detector.
    \item We demonstrate that the proposed method results in improved OD performance even when reducing the number of optical elements in a given lens stack.
\end{itemize}

In addition, we release our \href{https://github.com/princeton-computational-imaging/joint-lens-design}{\color{magenta}{code and designs}}\footnote{\url{https://github.com/princeton-computational-imaging/joint-lens-design}} in the hope of enabling further joint design applications.

\paragraph{Limitations}
In end-to-end optics design, the inherent resolution of the dataset used to represent real-world scenes---a result of the pixel count, imaging quality, and compression artifacts---needs to be discernibly superior to the modeled optics if meaningful conclusions are to be drawn.
Hence, we focus on simple lenses with strong geometrical aberrations, namely refractive lenses with two to four spherical elements whose combination of aperture and field of view (FOV) exceeds the capabilities of the lens configuration.
Incidentally, our method does not completely alleviate the need for human supervision; as in most lens design problems, a suitable lens design starting point is required for best performance.

\begin{table}[t]
    \setlength{\tabcolsep}{0em}
    \centering
    \footnotesize
    \input{related_work.tex}
    \caption{
        Comparison of related work on the joint optimization of refractive compound optics, where each criterion is fully~\greencheck, partially~\yellowcheck, or not~\redcheck met.
        See text for explanations.
        }
    \label{tab:related_work}
\end{table}

\begin{figure*}[t]
    \centering
    \includegraphics{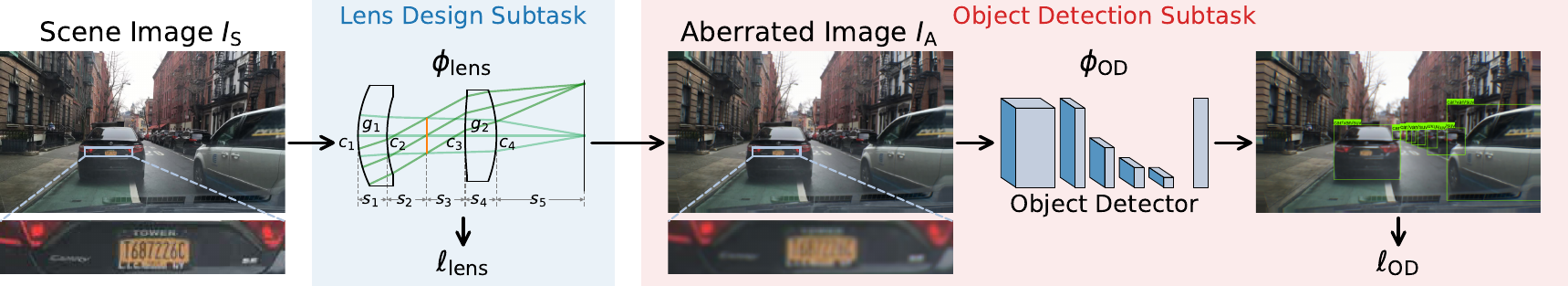}
    \caption{
        Overview of our joint design approach.
        To tackle the lens design and OD subtasks jointly rather than in isolation, we simulate realistic aberrations on the dataset images before they enter the object detector.
        The lens parameters~$\phi_\mathrm{lens}$ (with curvatures~$c$, spacings~$s$, and glass variables~$g$) are optimized both on a loss~$\ell_\mathrm{lens}$, which targets optical performance and geometrical constraints via exact ray tracing, and on the OD loss~$\ell_\mathrm{OD}$.
        The object detector parameters~$\phi_\mathrm{OD}$ are trained concurrently to minimize~$\ell_\mathrm{OD}$ while adjusting to the lens aberrations.
        }
    \label{fig:overview}
\end{figure*}

\section{Related Work}

We briefly review the existing literature on joint optics design based on three aspects: optics simulation, optimization, and integration with the downstream task.
\cref{tab:related_work} compares our work to other approaches that focus on compound optics.

\paragraph{Optics Simulation}
Many optics simulation models consist primarily in convolving the point spread function (PSF) of the optics design over the target image.
This approach is used in several works, in particular with DOEs~\cite{chang2019deep,metzler2020deep,sitzmann2018end,sun2020learning} that have a single surface where the PSF can be approximated using paraxial Fourier-based models.

In contrast, compound lenses have multiple surfaces with varying materials and surface profiles. 
To capture these optical systems accurately, exact ray tracing based on Snell's Law is typically used to complement paraxial optics.
As many existing models~\cite{kolb1995realistic,maeda2005integrating,steinert2011general,hanika2014efficient} are not end-to-end differentiable, recent works have introduced new differentiable lens models to enable the joint design of compound lenses.
\citet{tseng2021differentiable} employ a proxy model that learns the mapping between lens variables and PSFs using pre-generated lens data; while bypassing the intricacies of ray-based rendering, it adds a cumbersome training step that needs to be repeated for every lens configuration and only works for predefined variable boundaries.
\citet{sun2021end} apply Monte Carlo ray tracing from every pixel of the virtual detector at the expense of computational efficiency.
\citet{hale2021end} and \citet{li2021end} apply ray-tracing operations in a way that is reminiscent of conventional optical ray tracing~\cite{yabe2018optimization} to compute the PSFs, with the former assuming a Gaussian shape for the PSFs and the latter assuming a square entrance pupil.
The lens simulation model developed in our work is most similar to \citet{li2021end} with notable additions: we implement accurate ray aiming and avoid discontinuity artifacts by addressing distortion and relative illumination in separate steps (see~\cref{tab:related_work}).
Crucially and as opposed to our work, \cite{li2021end} does not optimize for computer vision tasks.
We note that our work deliberately focuses on spherical lenses to enable low-cost automotive OD.

\paragraph{Optics Optimization}
Conventional lens design tasks usually seek a design of suitable complexity that fulfills a given list of specifications; these are translated into a loss function that targets optical performance criteria as well as many manufacturing constraints~\cite{bentley2012field}.
The lens configuration is chosen to provide sufficient degrees of freedom (DOF) and dictates the number and nature of lens variables, notably: glass materials, spacings between each optical surface, and surface profile parameters---characterized by the curvature and, for aspherics, additional polynomial coefficients~\cite{schuhmann2019description}.
The de-facto optimizer for lens optimization is the Levenberg-Marquardt algorithm~\cite{girard1958excerpt,wynne1959lens}, which is also the default option~\cite{yabe2018optimization} in common optical design software~\cite{synopsys2018code, zemax2019user}.

In contrast, joint optics design optimizes the optics alongside the downstream neural network parameters using SGD-based optimization; as such, developing a lens optimization strategy that synergizes with SGD is a focus of our work, enabling end-to-end optimization for vision downstream tasks.
Conversely, previous works circumvent these difficulties by optimizing only a subset of both spacings and surface profile parameters~\cite{sun2021end,hale2021end,li2021end} or, in the case of~\cite{tseng2021differentiable}, limiting the variables within predefined boundaries (see \cref{tab:related_work}).

\paragraph{Joint Domain-Specific Optics Optimization}
Many downstream tasks can be grouped under the umbrella term of image reconstruction, where the goal is to retrieve the original image despite lens aberrations~\cite{li2021end,tseng2021neural} or environmental changes such as low-light imaging~\cite{tseng2021differentiable}.
This includes high-dynamic-range~\cite{sun2020learning,metzler2020deep}, large field-of-view~\cite{peng2019learned}, extended depth-of-field~\cite{sitzmann2018end}, and super-resolution~\cite{sitzmann2018end} imaging.
End-to-end design has also been applied to traditional vision tasks such as image classification~\cite{chang2018hybrid}, monocular depth estimation~\cite{chang2019deep, haim2018depth}, or OD~\cite{chang2019deep,tseng2021differentiable}.
While our approach supports any downstream task that can be trained with SGD optimization, here we focus on automotive OD, relevant to self-driving vehicles and autonomous robots.

\begin{figure*}[tbp]
    \centering \includegraphics{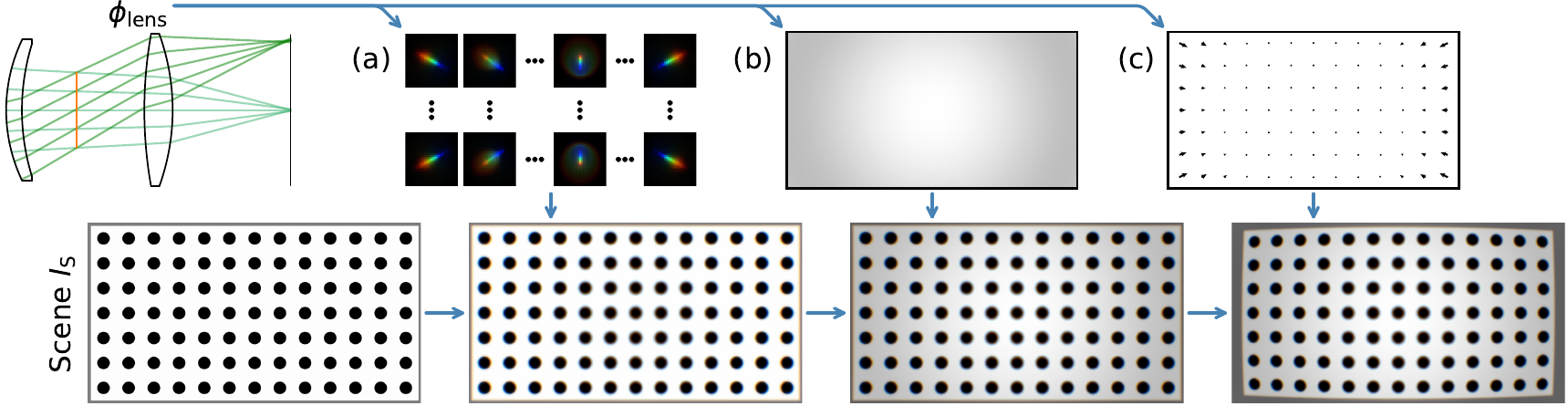}
    \caption{
        From the lens parameters~$\phi_\mathrm{lens}$, our lens simulation model employs exact differentiable ray tracing to compute the spatially varying PSF grid~(a), relative illumination map~(b), and distortion field~(c), which are successively applied to the scene image~$I_\mathrm{S}$ to simulate realistic geometrical aberrations (relative illumination is~20$\times$ amplified for clarity).
        }
    \label{fig:lens_simulation}
\end{figure*}

\section{Differentiable Compound Lenses}

In this work, we first introduce a method for the end-to-end modeling and optimization of compound lenses in computer vision tasks.
We then apply the proposed method to OD as illustrated in \cref{fig:overview}.
We use natural images as input to our method and as approximations of real-world scenes with the following underlying assumptions (see supp.): the objects are infinitely distant, the RGB values are proportional to the luminance, and the FOV of the simulated lens matches the scene.
These approximations allow us to rely on existing image datasets to study the effect of strong geometrical aberrations such as the ones of poorly corrected optics, that is, lenses without the required DOF to correct the aberrations under the desired specifications.
Indeed, sophisticated lenses (\eg, with a larger number of elements) are of limited interest in our work since they do not significantly impact image quality.
Therefore, we focus on simple lenses composed of a few (2--4) spherical lens elements while noting that poorly corrected optics are \emph{harder} to simulate accurately due to larger PSFs, distortion, and spatial variations.

The throughput of a lens is an important consideration in low-light OD.
As such, we design all lenses to have similar and relatively high throughput with a fixed f-number and no any optical vignetting.
Incidentally, we fix both the FOV and focal length~$f$ such that $f = \nicefrac{d}{2\tan(\mathrm{FOV}/2)}$ to ensure that the corners of the virtual image sensor (with diagonal~$d$) correspond to the maximum FOV---assuming reasonable distortion and defocus.

In the following, we describe the core components of the proposed method.
In \cref{sec:simulation}, we elaborate on our complete differentiable lens simulation model.
In \cref{sec:optimization}, we detail the lens parameters $\phi_\mathrm{lens}$ and our joint optimization strategy.
We assess the proposed method experimentally in \cref{sec:experiments}.

\section{Optical Image Formation Model}
\label{sec:simulation}

Sampling and tracing rays from every pixel of the virtual detector~\cite{kolb1995realistic,sun2021end} is computationally prohibitive for our joint design approach.
Instead, our differentiable lens simulation model applies a spatially varying convolution to the scene image~$I_\mathrm{S}$ to generate the aberrated image
\begin{equation}
    I_\mathrm{A}(x', y') \approx
    \mathrm{PSF}(x', y') * I_\mathrm{S}(x', y')
    \,.
    \label{eq:conv}
\end{equation}
We model \cref{eq:conv} by discretizing the image into a grid of patches that are convolved with their corresponding PSF.
While the PSFs can theoretically take distortion and relative illumination into account (as in \citet{tseng2021differentiable}), here we simulate them in separate steps as shown in \cref{fig:lens_simulation}.
This avoids discontinuity artifacts and, in the case of distortion, artificial blurring~\cite{maeda2005integrating} as well as an increased computational burden caused by large uncentered PSFs.

\begin{figure}[t]
    \centering
    \includegraphics{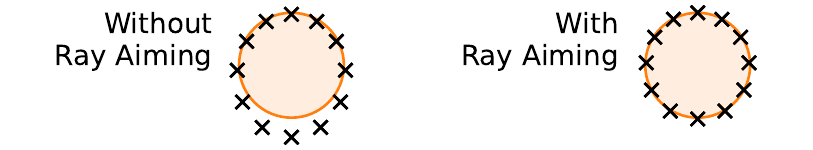}
    \caption{
        Rays that are targeted at the outer edge of the aperture stop (cross section is shown in orange) successfully hit the target area when using the proposed ray-aiming correction step, and badly miss otherwise.
        Illustrated here is for the f/2 Tessar lens used in \cref{sec:experiments} at full field of view (\ang{25}, see supp.\ for more examples).
        }
    \label{fig:ray_aiming}
\end{figure}

\paragraph{Ray Tracing}
Exact ray tracing is achieved by alternating between two operations: 1) updating the coordinates of the rays from one interface to the next, and 2) updating the direction cosines following Snell's Law.
In practice, we batch the operations over $n_\mathrm{r}=n_\mathrm{h}n_\mathrm{w}n_\mathrm{p}$ rays, where $n_\mathrm{h}$, $n_\mathrm{w}$, and $n_\mathrm{p}$ are the number of field values, wavelengths, and pupil coordinates.
All rays are initialized at the entrance pupil.
Unlike~\cite{li2021end}, we introduce a ray-aiming correction step which is critical to accurately simulate lenses with strong pupil aberrations (see \cref{fig:ray_aiming}); as in~\cite{cote2021deep}, the initial transverse ray coordinates are scaled by deforming the entrance pupil into a field-dependent elliptic shape (see supp.).

\begin{figure}[tbp]
    \centering \includegraphics{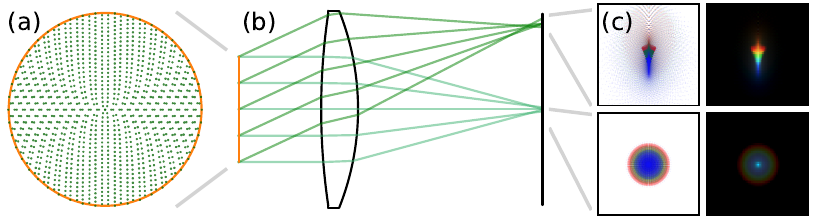}
    \caption{
        Computation of the geometrical PSFs used to simulate realistic aberrations.
        We first initialize rays at the entrance pupil~(a), which in this example overlaps with the aperture stop (orange) located in object space.
        We propagate the rays using exact ray-tracing operations~(b) to obtain the spot diagrams.
        Then, we apply kernel density estimation to retrieve the PSFs for all field values~(c).
        }
    \label{fig:psf}
\end{figure}

\paragraph{Geometrical PSFs}
Under dominant geometrical aberrations, diffraction can be safely ignored and the PSFs can be computed through the ray-counting method: setting a virtual detector on the image plane and counting the rays hitting each bin (see \cref{fig:psf}).
As the PSFs of axially symmetric lenses are invariant to azimuth, here they are sampled radially at $n_\mathrm{h} = 21$~equidistant field values~$h$, then interpolated, rotated, and resized to fill the PSF grid (see \cref{fig:lens_simulation}(a)).

First, for each field, we span the entrance pupil uniformly with rays---each representing an equal pupil area and amount of energy---and trace them up to the image plane to obtain the $x \in \mathbb{R}^{n_\mathrm{r}}$ and $y \in \mathbb{R}^{n_\mathrm{r}}$ coordinates that compose the spot diagrams.
The pupil sampling scheme ($n_\mathrm{p} = 2048$) corresponds to 32~equally spaced concentric circles with jittering to properly sample the outer edge of the pupil (see \cref{fig:psf}(a)).
We trace $n_\mathrm{w} = 15$~wavelengths: 5~for each color channel which are selected from the quantum efficiency of a typical sensor (here, we use the Sony IMX172, see supp.).

Then, we center a square virtual grid for each field~$h$ at the spot diagram centroid~$\overline{y_h} = (\nicefrac{1}{n_\mathrm{w}n_\mathrm{p}})\sum_{w,p} y_{h,w,p}$.
We set the size of the virtual grid to \qty{260}{\um} as to collect all rays throughout the full optimization process, and split it into \numproduct{65x65} bins.
Instead of naive ray counting, we employ the differentiable alternative of kernel density estimation (KDE) using a Gaussian kernel with a bandwidth half the size of a bin, which effectively spreads the energy of each ray over multiple bins.
Incidentally, we reduce the computational burden by duplicating all rays and bins across the $y$-axis.

\paragraph{Spatially Varying Convolution}
We employ the spatially varying overlap-add method~\cite{hirsch2010efficient} using \numproduct{9x9} rectangular image patches with corresponding PSFs (see \cref{fig:lens_simulation}(a)).
Each PSF in the grid is a weighted average of the sampled PSFs---the weight for a field~$h$ corresponds to the proportion of the patch that is closest to it---that is rotated to the appropriate angle, then rescaled according to the image resolution.
In contrast to naive interpolation, 
this weighted average scheme involves the full FOV of the lens in the simulation and optimization pipeline.
For smooth interpolation, we use a 2D Hann window with \qty{25}{\percent} overlap.

\paragraph{Relative Illumination}
Assuming elliptic pupils, we can obtain a coarse approximation of the relative illumination factor~$R_h$ at a given field of interest~$h$ from the direction cosines of two meridional rays and one sagittal ray~\cite{rimmer1986relative}.
We apply the operation monochromatically (\qty{587.6}{\nm}), then interpolate the values according to the radial coordinate of each pixel.
Finally, the aberrated image is pixel-wise multiplied with the relative illumination map (see~\cref{fig:lens_simulation}(b)).

\paragraph{Distortion}
To efficiently simulate distortion, we approximate the relative distortion shift $D_h$ at each field $h$ by comparing the mean ray height $\overline{y_h}$ at the image plane to the undistorted reference value $y_{h, \mathrm{ref}}$
\begin{equation}
    D_h = 
    \frac{\overline{y_h} - y_{h, \mathrm{ref}}}{y_{n_\mathrm{h}, \mathrm{ref}}}
    \,,
    \label{eq:distortion}
\end{equation}
where the reference values~$y_{h, \mathrm{ref}}$ are the result of a monochromatic paraxial ray-tracing operation (\qty{587.6}{\nm}).
Next, the distorted $(x', y')$ coordinates are computed by linearly interpolating and rotating the distortion shift based on the field position of each pixel.
Finally, the image is warped using bicubic interpolation (see \cref{fig:lens_simulation}(c)).

\section{Joint Optimization}
\label{sec:optimization}

Given the differentiable image formation model from \cref{sec:simulation}, we now seek to freely optimize all lens variables on downstream tasks without compromising manufacturability, which can be facilitated by employing well-defined constraints.
We note, however, that there exists no universal set of rules to assess whether a lens design is manufacturable; it notably depends on the expertise and equipment at hand.

\paragraph{Lens Variables}
We consider a compound lens as a stack of~$M$ spherical glass elements with~$K$ interfaces (including the aperture stop, but excluding the image plane) where neighboring lens elements are either air spaced or cemented together.
Lens variables are denoted $\phi_\mathrm{lens} = \left(c', s', g\right)$, where $c'\in\mathbb{R}^{K-2}$ are normalized curvatures of the spherical interfaces, $s'\in\mathbb{R}^K$ normalized glass and air spacings, and $g\in\mathbb{R}^{M\times d_\mathrm{glass}}$ sets of $d_\mathrm{glass}$ glass variables representing the dispersion curve of each glass element.
The last curvature (before the image plane) is not optimized, but algebraically solved at every training iteration to enforce a unit focal length $f' = 1$.
Then, the curvatures~$c$ and spacings~$s$ are obtained by scaling their normalized counterparts to the desired focal length: $c = \nicefrac{c'}{f}; s = s' f$.
As in~\cite{cote2022inferring}, we implement a \emph{paraxial image solve} to help the lens remain mostly in focus throughout optimization, which requires computing the back focal length (BFL) to locate the paraxial image plane with respect to the last optical surface.
Then, the last airspace $s_K = s_K' f + \mathrm{BFL}$ is retrieved from the normalized defocus~$s_K'$, which acts as the optimized variable.

\begin{figure}[tbp]
    \centering
    \includegraphics{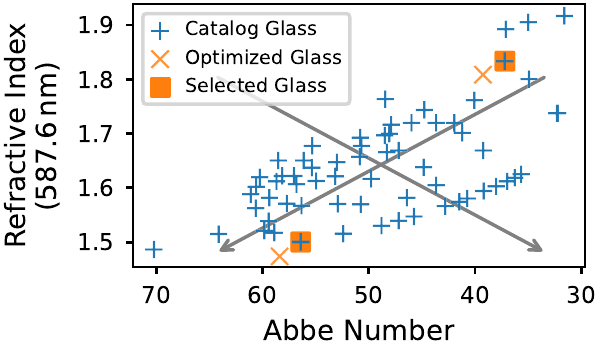}
    \caption{
        Quantized continuous glass variables.
        Using continuous glass variables enables the exploration of the solution space, but the actual variables employed to model the lens are always their closest catalog counterparts ($n_\mathrm{cat} = 65$).
        The arrows indicate the whitened coordinate system.
        }
    \label{fig:glass_model}
\end{figure}

\paragraph{Glass Variables}
Our aim in optimizing glass variables is to find the best set of materials among the \textit{catalog glasses} $\left(g'_1, g'_2, \ldots, g'_{n_\mathrm{cat}}\right)$.
To this end, we consider $n_\mathrm{cat} = 65$~recommended glasses from the Ohara catalog~\cite{corporation2019optical2} (see \cref{fig:glass_model}).
We model the dispersion curve of each glass material with $d_\mathrm{glass} = 2$ variables: the refractive index at the ``d" Fraunhofer line (\qty{587.6}{\nm}) and the Abbe number.
As in~\cite{sun2021end}, we use the approximate dispersion model $n(\lambda) \approx A + B/\lambda^2$ to retrieve the refractive index at any wavelength~$\lambda$, where $A$ and $B$ follow from the definition of the ``d"-line refractive index and Abbe number.
We obtain our normalized glass variables~$g$ by fitting a whitening transformation on the refractive indices and Abbe numbers of all catalog glasses.

\paragraph{Quantized Continuous Glass Variables}
Using continuous relaxations for glass optimization presents several issues in SGD-based end-to-end optimization.
Requiring the glass variables to converge to catalog glasses while allowing them to vary significantly during training is challenging as it would require the delicate tuning of scheduled constraints.

To avoid this issue, we introduce \textit{quantized continuous glass variables}: glass variables that only exist in discrete sets, but retain the optimizable property of continuous variables.
As illustrated in~\cref{fig:glass_model}, in the forward pass, we replace each set of variables~$g_m$ with its closest catalog glass counterpart
\begin{equation}
    g_m^* = 
    \mathop{\arg\min}_j
    ||g_m - g'_\mathrm{j}||^2_2
    \,.
\end{equation}
As this operation is not differentiable, we approximate its gradient using the ``gradient step-through" operator~\cite{van2017neural}.
This operation allows glass variables to undergo meaningful optimization while ensuring that they always match available glass materials.
As our approach allows large jumps in lens performance when new catalog glasses are selected from one optimization step to the other, we couple it with a \textit{glass variable loss}~$\ell_\mathrm{GV}$ to help the free variables stick close to the selected glasses, therefore limiting the magnitude and frequency of such jumps.
The loss minimizes the squared distance between each set of continuous glass variables~$g_m$ and the closest catalog glass
\begin{equation}
    \ell_\mathrm{GV} = 
    \sum_m
    ||g_m - g_m^*||^2_2
    \,.
\end{equation}
We find empirically that the lenses optimized with this approach retrieve a good performance within a few steps.

\paragraph{Design Losses}
In practice, we find that the training signal due to the downstream OD loss is often noisy and can make reliable optimization challenging.

To account for this as well as several manufacturing constraints, we add a set of design losses to assist the optimization.
First, we complement the noisy downstream detection loss with a \textit{spot size loss}~$\ell_\mathrm{S}$ for stability.
The spot size is equivalent to the RMS size of the PSF (for a given field~$h$) and is computed from the same transversal ray coordinates~$x$ and~$y$ that compose the spot diagram (see \cref{sec:simulation}).
We formulate~$\ell_\mathrm{S}$ as the average spot size across all field values
\begin{equation}
    \ell_\mathrm{S} =
    \frac{1}{n_\mathrm{h}}
    \sum_h
    \sqrt{
    \frac{1}{n_\mathrm{w}n_\mathrm{p}}
    \sum_{w, p} 
    \left(y_{h, w, p} - \overline{y_h}\right)^2
    +
    x_{h, w, p}^2
    }
    ~.
    \label{eq:spot_loss}
\end{equation}
Next, we add two additional \textit{ray path} and \textit{ray angle} losses, which are defined by reusing intermediate operands from every ray~$r$ involved in the computation of the spot diagrams or spot size.
The \textit{ray path loss}~$\ell_\mathrm{RP}$ avoids overlapping surfaces, enforces sufficient center/edge thicknesses in glass elements, and imposes a sufficient image clearance (the clear space between the last element and the image sensor).
It is defined using the horizontal distance $\Delta z \in \mathbb{R}^{K\times n_\mathrm{r}}$ traveled by every ray~$r$ across every glass or air spacing~$k$.
We want all rays to travel a horizontal distance bounded between a lower threshold $\Delta z_\mathrm{min}^{(k)}$ and an upper threshold $\Delta z_\mathrm{max}^{(k)}$ that depend on the nature of the spacing.
In our experiments, these are set to enforce a minimum distance of \qty{0.01}{\mm} in airspaces and \qty{12}{\mm} for image clearance, and a distance between 1--3\,\si{\mm} in glass.
The loss is formulated as
\begin{align}
    \ell_\mathrm{RP} &= \frac{1}{n_\mathrm{r}}\sum_{k, r} \max 
    \left(
        \Delta z_\mathrm{min}^{(k)} - \Delta z_{k, r}, 0
    \right)
    \nonumber
    \\
    & +
    \max 
    \left(
        \Delta z_{k, r} - \Delta z_\mathrm{max}^{(k)}, 0
    \right)
    \,.
    \label{eq:ray_path}
\end{align}
The \textit{ray angle loss}~$\ell_\mathrm{RA}$ limits all angles of incidence~$\theta$ and refraction~$\theta'$ to a threshold $\theta_\mathrm{max} = \ang{60}$; this aims to avoid ray failures, stabilize the optimization process, and improve tolerancing.
Tracing rays through spherical surfaces involves the computation of intermediate values $\zeta = \cos^2\left(\theta\right)$ and $\zeta' = \cos^2\left(\theta'\right)$, where negative values imply ray-tracing failure in the form of missed surfaces for $\zeta$ and total internal reflection for $\zeta'$.
Similar to~$\ell_\mathrm{RP}$ (\cref{eq:ray_path}), this loss is based on intermediate ray-tracing operands occurring at every interface $k$ prior to the image plane
\begin{align}
    \ell_\mathrm{RA} &= \frac{1}{n_\mathrm{r}}\sum_{k, r} \max 
    \left(
        \cos^2\left(\theta_\mathrm{max}\right) - \zeta_{k, r}, 0
    \right)
    \nonumber
    \\
    & +
    \max
    \left(
        \cos^2\left(\theta_\mathrm{max}\right) - \zeta'_{k, r}, 0
    \right)
    \,.
\end{align}

\begin{figure*}[t]
    \centering
    \includegraphics{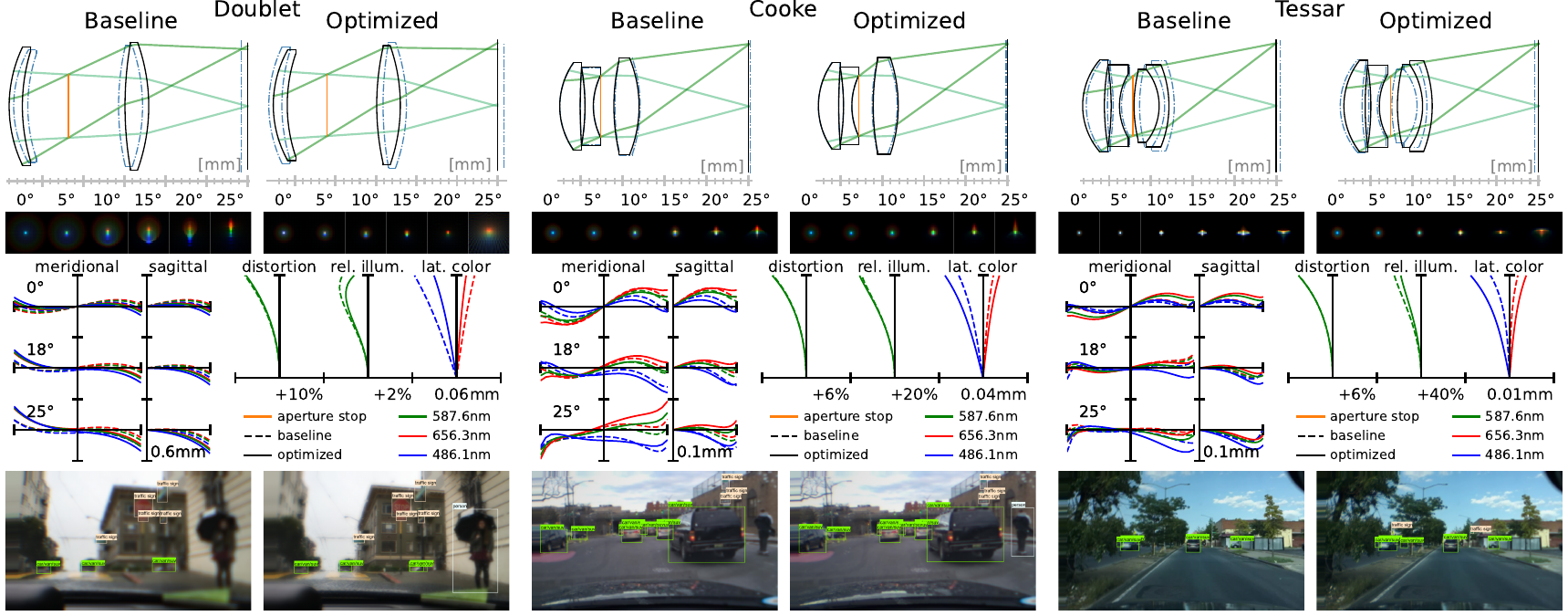}
    \caption{
        Comparison between lenses optimized for spot size (Baseline) or object detection (Optimized) under 2\texttimes\ simulated resolution, for 2- (Doublet), 3- (Cooke), and 4-element (Tessar) designs.
        From top to bottom, we show 1) the lens designs (dashed lines represent the baseline/optimized counterpart); 2) PSFs for different fields; 3) aberration charts (left: ray fan plots; right: field curves); and 4) images where the bottom-right corner corresponds to on-axis imaging (\ang{0}) and the top-left corner to the maximum field of view (\ang{25}).
        Here, in contrast to the baseline lenses, the optimized lenses detect additional persons (Doublet and Cooke) or traffic signs (Tessar).
        }
    \label{fig:qualitative}
\end{figure*}

\paragraph{End-to-End Optimization}
All previous loss terms are combined to define a loss~$\ell_\mathrm{lens}$ that operates exclusively on the lens design parameters~$\phi_\mathrm{lens}$
\begin{equation}
    \ell_\mathrm{lens} =
    \ell_\mathrm{S} +
    \lambda_\mathrm{RP} \ell_\mathrm{RP} +
    \lambda_\mathrm{RA} \ell_\mathrm{RA} +
    \lambda_\mathrm{GV} \ell_\mathrm{GV}
    \,,
    \label{eq:lens_design_loss}
\end{equation}
where we set $\lambda_\mathrm{RP}=100$, $\lambda_\mathrm{RA}=100$, and $\lambda_\mathrm{GV}=0.01$.
\Cref{eq:lens_design_loss} can be used in isolation to optimize the baseline lenses for spot size, but is also combined with object detection losses $\ell_\mathrm{OD}$ to define the joint loss
\begin{equation}
    \ell_\mathrm{joint} =
    \ell_\mathrm{OD} +
    \lambda_\mathrm{lens} \ell_\mathrm{lens}
    \,,
    \label{eq:joint_loss}
\end{equation}
where $\lambda_\mathrm{lens}$ is set individually for each lens.

\section{Experiments}
\label{sec:experiments}

In this section, we validate the proposed method on a variety of different lens design tasks, compare it to existing design methods, and confirm the effectiveness of components of the method in ablation experiments.
To this end, we first introduce the dataset and training approach employed for all experiments, describe baseline lens designs in~\cref{sec:lenses}, and then discuss OD design experiments in~\cref{sec:experiments_baseline}.

\paragraph{Datasets}
We conduct our experiments with the BDD100K dataset~\cite{yu2020bdd100k} containing 80k (70k/10k for training/evaluation) all-in-focus images with moderate resolution (\numproduct{1280x720}) and minimal visible aberrations even before downsampling, which makes it suitable for our experiments.
We consider six aggregated classes (car/van/suv, bus/truck/tram, bike, person, traffic light, and traffic sign).
We also evaluate our trained models on the Udacity autonomous driving dataset~\cite{udacity2022annotated} which contains 14k higher-resolution images (\numproduct{1920x1200}) that were annotated using the same six classes.

\paragraph{Sensor Simulation}
We consider a sensor diagonal $d = \qty{16}{\mm}$ and a quantum efficiency curve that follows the Sony IMX172 sensor for representative wavelength sampling (see supp.).
The lens model is applied to the unaltered dataset images which are subsequently resized (\numproduct{1024x1024}) and passed to the OD model.
To observe larger OD performance degradations, we also simulate a 2\texttimes-increased resolution in which the aberrations appear proportionally larger; in this setting, the dataset image occupies only the upper-left quadrant of the original virtual scene, and we simulate the aberrations accordingly as shown in \cref{fig:qualitative}.

\paragraph{Detector and Training Methodology}
We use the RetinaNet~\cite{lin2017focal} object detector with a ResNet-50 backbone~\cite{he2016deep} for all experiments.
We train all OD models with a batch size of 8 and we jointly optimize the lens and OD model with Adam~\cite{kingma2015adam}. 
The learning rates are set to \qty{5e-5} for $\phi_\mathrm{OD}$ and \qty{5e-3} for $\phi_\mathrm{lens}$ over 50k~steps, then both are decayed to 0 over 100k~subsequent steps following a half cosine cycle.

\paragraph{Lens Distortion in Object Detection}
The object-matching operation commonly used in IoU losses interferes with distortion since it moves the content associated with predefined anchors.
To account for this, we use \cref{eq:distortion} to apply a correction step to all ground truth boxes when computing the OD losses, by shifting the midpoint of each bounding box segment, then drawing a new bounding box around the shifted coordinates.
However, to evaluate the average precision (AP) of the OD models in an unbiased manner, we apply the correction step to the \emph{predicted} boxes instead.

\subsection{Baseline Lenses}
\label{sec:lenses}

We conduct our experiments using typical lenses with 2--4 elements as visualized in \cref{fig:qualitative}.
In contrast to the 2-element Doublet, the 3-element Cooke triplet lens has sufficient DOF for moderate aperture and FOV imaging~\cite{smith2004modern}.
The 4-element Tessar lens can be seen as a modified Cooke triplet with more DOF~\cite{smith2004modern}.
All lenses are optimized for the same first-order specifications, namely f/2 for aperture, $\pm\ang{25}$ for FOV, and focal length~$f = \qty{17.2}{\mm}$.
We note that even 4-element spherical lenses do not have the required DOF to adequately correct geometrical aberrations under this combination of aperture and FOV~\cite{smith2004modern}.
Therefore, our results can be interpreted as approximate lower bounds on OD performance; reducing the aperture or FOV would likely lead to similar or better performance in all cases.

To obtain our baseline lenses, we follow common practice and start from several starting points with various configurations---namely, different aperture stop locations or cemented interfaces---from varied sources~\cite{smith2004modern,cote2021lensnet,corporation2019optical}, reoptimize each of them using \cref{eq:lens_design_loss}, and select the ones that have the best average spot size according to \cref{eq:spot_loss}.

\subsection{Automotive Object Detection}
\label{sec:experiments_baseline}

We report our lens designs optimized for object detection in \cref{tab:results} in terms of mean average precision (AP)---averaged over the IoU thresholds (0.5, 0.55, \ldots, 0.95) and all six object classes.
Additionally, we report the averaged PSNR and SSIM image quality metrics to compare the images before and after simulating the blur-inducing aberrations (prior to applying relative illumination and distortion).

\begin{table}[tbp]
    \centering
    \setlength{\tabcolsep}{.25em}
    \footnotesize
    \setlength{\aboverulesep}{1pt}
    \setlength{\belowrulesep}{1pt}
    \renewcommand*{\arraystretch}{0.75}
    \input{results.tex}
    \caption{
        Mean spot size (\si{\um}, see \cref{eq:spot_loss}), image quality metrics (PSNR and SSIM), and final OD performance~(AP) across varied experimental settings.
        The lens and OD model parameters $\phi_\mathrm{lens}$ and $\phi_\mathrm{OD}$ are either optimized~\unlock or fixed~\lock.
        When $\phi_\mathrm{OD}$ is fixed, we use the same parameters as in the perfect optics baseline (first row).
        Settings with~\redstar are visualized in \cref{fig:qualitative} (see supp. for others).
    }
    \label{tab:results}
\end{table}

To provide approximate upper bounds for OD performance, which is equivalent to training and evaluating the OD models without any aberrations, we first report the AP under ``perfect" optics (first row of \cref{tab:results}).
We also evaluate this trained model when simulating the effect of each baseline lens; this scenario, labeled \lock--\lock in \cref{tab:results}, is akin to attempting OD on strongly aberrated images using an off-the-shelf OD model.
This leads to a large decrease in AP (\mbox{-19.5}/\mbox{-7.4}/\mbox{-2.9} for 2/3/4 elements on 1\texttimes\ res.\ on BDD).

Then, we fine-tune the OD model to account for the simulated aberrations (\lock--\unlock), by modeling the lens using its known design.
In practice, this could also be achieved by capturing a dataset using the manufactured lens and training on it.
This greatly alleviates the AP drop compared to ``perfect'' optics (\mbox{-3.8}/\mbox{-1.1}/\mbox{-0.7} for 2/3/4 elements on 1\texttimes\ res.\ on BDD) despite the significant degradation in image quality.

Finally, we jointly optimize the lens alongside the OD model (\unlock--\unlock) using \cref{eq:joint_loss}, which results in an increase in AP over the scenario \lock--\unlock (\eg, \mbox{+3.1}/\mbox{+0.2}/\mbox{+0.9} for 2/3/4 elements on 2\texttimes\ res.\ on BDD), validating the benefit of joint optics/OD optimization.
On 1\texttimes-res.\ BDD, the joint optimization allows the Cooke triplet (AP of 33.3, up from 33.0) to nearly reach the performance of the baseline Tessar lens (AP of 33.4) which has one additional lens element.
The improvements are not specific to the BDD100K dataset: our experiments validate that they generalize to the higher-resolution Udacity dataset.
We note that the optimized lenses improve OD performance despite a mean spot size similar to or worse than the baseline lenses (see supp. for a tolerancing analysis and a comparison to other ray-tracing approaches).

\begin{table}[t]
    \centering
    \footnotesize
    \setlength{\tabcolsep}{0em}
    \begin{tabularx}{\linewidth}{cXcXcXcXc}
        \toprule
        Setting && Optics && Spot (\si{\um})$_\downarrow$ && Vig. rays$_\downarrow$ && AP$_\uparrow$ \\
        \midrule
        PM~\cite{tseng2021differentiable} && \multirow{2}*{Tessar (1\texttimes\ res.)} && 176.4 && \qty{13.2}{\percent} && 30.8 (PM); 29.0 (RT) \\
        \textbf{Ours} &&&& 14.8 && \qty{0.0}{\percent} && \textbf{33.6} (RT) \\
        \midrule
        PM~\cite{tseng2021differentiable} && \multirow{2}*{Tessar (2\texttimes\ res.)} && 104.0 && \qty{7.4}{\percent} && 23.9 (PM); 18.4 (RT) \\
        \textbf{Ours} &&&& 24.7 && \qty{0.0}{\percent} && \textbf{32.2} (RT) \\
        \bottomrule
    \end{tabularx}
    \caption{
        Comparison with the proxy model (PM) of \citet{tseng2021differentiable} on the joint optimization of the Tessar lens.
        We report the AP on BDD100K, where aberrations are modeled using either the PM or exact ray tracing (RT).
        We also report the mean spot size and proportion of vignetted rays, where \qty{0}{\percent} indicates that the design specifications (f-number, FOV, and no vignetting) are fulfilled.
        }
    \label{tab:comparison}
\end{table}

\paragraph{Comparison to Proxy Model}
In \cref{tab:comparison}, we perform the Tessar lens experiments using a proxy model (PM) from \citet{tseng2021differentiable}.
We train the PM on the data of 10k variations of the baseline Tessar lens (see supp.) to output the PSF, relative illumination factor~$R_h$, and distortion shift~$D_h$ for a given field value~$h$ and set of 22~lens variables.
As in~\cite{tseng2021differentiable}, we predetermine boundaries for all lens variables, which we optimize only on $\ell_\mathrm{OD}$ (\ie, $\lambda_\mathrm{lens} = 0$).
Compared to our method, the experiments show a significant decrease in AP even when the aberrations are modeled using the PM rather than exact ray tracing during evaluation.
Moreover, the lenses optimized using the PM do not fulfill the design specifications as a significant proportion of rays are vignetted, further validating the proposed method.

\begin{table}[t]
    \centering
    \footnotesize
    \setlength{\tabcolsep}{0em}
    \begin{tabularx}{\linewidth}{cXcXc}
        \toprule
        Setting && AP$_\uparrow$ && Comment \\
        \midrule
        \textbf{Complete methodology} && \textbf{32.2} && -- \\
        Continuous glass variables && 25.9 && Unrealistic glass ($\ell_\mathrm{GV} = 0.91$) \\
        No paraxial image solve && 26.5 && Last airspace is $s_K = s_K' f$ \\
        No ray path loss $\lambda_\mathrm{RP} = 0$ && 15.4 && Unfeasible design (ray failures) \\
        No ray angle loss $\lambda_\mathrm{RA} = 0$ && 24.2 && Unfeasible design (ray failures) \\
        No spot size loss $\lambda_\mathrm{S} = 0$ && 11.1 && Spot size of 142 um ($\uparrow$9.6\texttimes) \\
        \bottomrule
    \end{tabularx}
    \caption{
        Ablation study on the joint optimization of the Tessar lens (2\texttimes\ res.), where we report the AP on BDD100K.
        }
    \label{tab:ablation}
\end{table}

\paragraph{Ablations}
In \cref{tab:ablation}, we report ablation experiments on the joint design of the Tessar lens for 2\texttimes\ resolution.
The experiments validate that each component of the proposed method is required to avoid instability; in this setting, any component removal leads to a drop in OD performance and, in some cases, in manufacturability issues (see supp.).
In particular, using continuously relaxed glass variables not only leads to unrealistic glass materials but also adds instability that can result in poorly behaved designs.
This ablation experiment validates the role of glass material optimization in lens design for downstream detection tasks. 

\section{Conclusion}
Where previous works in joint optics design attempt to optimize compound lenses over only a subset of possible surface profiles and spacings, here we establish a novel differentiable lens model and optimization method to enable the free optimization of all lens variables; notably, \emph{quantized continuous glass variables} circumvent issues due to the categorical nature of glass materials.
On automotive~OD, we consistently observe \emph{improvements in detection even when reducing the number of elements} in a given lens stack.
Along with the release of code, we hope that this work will enable exciting future research directions such as combining different lens components (\eg, aspherics or diffractive optical elements), modeling scenes with high-resolution multispectral data, or enabling depth-sensitive downstream tasks.

\paragraph{Acknowledgments}
\textls[-5]{
    This research was supported by the Sentinel North program of Université Laval and NSERC.
    Felix Heide was supported by an NSF CAREER Award (2047359), a Packard Foundation Fellowship, a Sloan Research Fellowship, a Sony Young Faculty Award, a Project X Innovation Award, and an Amazon Science Research Award.
}

%% file: related_work.tex
\begin{tabularx}{\linewidth}{m{0.25\linewidth}XXXXX}
    \toprule
    &
    {\footnotesize Tseng~\cite{tseng2021differentiable}} &
    {\footnotesize Sun~\cite{sun2021end}} &
    {\footnotesize Hale~\cite{hale2021end}} &
    {\footnotesize Li~\cite{li2021end}} &
    {\footnotesize Ours} \\
    \midrule
    \multicolumn{6}{l}{\textbf{Differentiable Lens Model}}\\ 
    \midrule
    Hands-Free & 
    \red & 
    \green & 
    \green & 
    \green & 
    \green \\
    Efficient & 
    \green & 
    \red & 
    \green & 
    \green & 
    \green \\
    Accurate PSFs & 
    \green & 
    \green & 
    \red & 
    \yellow & 
    \green \\
    Distortion & 
    \yellow & 
    \green & 
    \yellow & 
    \yellow & 
    \green \\
    Aspherics & 
    \green & 
    \green & 
    \yellow &
    \green & 
    \red \\
    \midrule
    \multicolumn{6}{l}{\textbf{Optimized Lens Variables}}\\ 
    \midrule
    No Boundaries & 
    \red & 
    \green & 
    \green & 
    \green & 
    \green \\
    Spacings & 
    \green & 
    \red & 
    \yellow & 
    \red & 
    \green \\
    Surface Profiles & 
    \green & 
    \yellow & 
    \red & 
    \yellow & 
    \green \\
    Glass Materials & 
    \red & 
    \red & 
    \red & 
    \red & 
    \green \\
    \bottomrule
\end{tabularx}

%% file: results.tex
\begin{tabularx}{\linewidth}{Xccccccc}
    \toprule
    \multirow{2}*{\textbf{Optics}} & \multirow{2}*{$\phi_\mathrm{lens}$} &  & \multicolumn{2}{c}{BDD100K} & \multirow{2}*{$\phi_\mathrm{OD}$} & BDD100K & Udacity \\
    \arrayrulecolor{lightgray}\cmidrule(lr{4pt}){4-5} \cmidrule(lr{4pt}){7-7} \cmidrule(lr{4pt}){8-8}\arrayrulecolor{black} &  & Spot$_\downarrow$ & PSNR$_\uparrow$ & SSIM$_\uparrow$ &  & AP$_\uparrow$ & AP$_\uparrow$ \\
    \midrule
    Perfect & -- & -- & -- & -- & -- & 34.1 & 28.2 \\
    \midrule
    \multirow{3}*{\shortstack{Doublet\\(1\texttimes\  res.)}} & \multirow{2}*{\lock} & \multirow{2}*{80.1} & \multirow{2}*{26.6} & \multirow{2}*{0.82} & \lock & 14.6 & 11.5 \\
    \arrayrulecolor{lightgray}\cmidrule(l{2pt}){6-8}\arrayrulecolor{black} &  &  &  &  & \unlock & 30.3 & 23.5 \\
    \arrayrulecolor{lightgray}\cmidrule(l{2pt}){2-8}\arrayrulecolor{black} & \unlock & 135.5 & 27.6 & 0.85 & \unlock & 32.0\color{Green}{(+1.7)} & 25.6\color{Green}{(+2.1)} \\
    \midrule
    \multirow{2}*{\shortstack{{\redstar}Doublet\\(2\texttimes\  res.)}} & \lock & 80.1 & 24.3 & 0.78 & \unlock & 25.0 & 18.4 \\
    \arrayrulecolor{lightgray}\cmidrule(l{2pt}){2-8}\arrayrulecolor{black} & \unlock & 124.5 & 25.5 & 0.81 & \unlock & 28.1\color{Green}{(+3.1)} & 21.9\color{Green}{(+3.5)} \\
    \midrule
    \multirow{3}*{\shortstack{Cooke\\(1\texttimes\  res.)}} & \multirow{2}*{\lock} & \multirow{2}*{30.6} & \multirow{2}*{29.1} & \multirow{2}*{0.88} & \lock & 26.7 & 20.6 \\
    \arrayrulecolor{lightgray}\cmidrule(l{2pt}){6-8}\arrayrulecolor{black} &  &  &  &  & \unlock & 33.0 & 26.3 \\
    \arrayrulecolor{lightgray}\cmidrule(l{2pt}){2-8}\arrayrulecolor{black} & \unlock & 36.6 & 28.7 & 0.87 & \unlock & 33.3\color{Green}{(+0.3)} & 26.8\color{Green}{(+0.5)} \\
    \midrule
    \multirow{2}*{\shortstack{{\redstar}Cooke\\(2\texttimes\  res.)}} & \lock & 30.6 & 27.2 & 0.84 & \unlock & 31.5 & 24.9 \\
    \arrayrulecolor{lightgray}\cmidrule(l{2pt}){2-8}\arrayrulecolor{black} & \unlock & 36.5 & 26.6 & 0.83 & \unlock & 31.7\color{Green}{(+0.2)} & 24.8\color{red}{(\textminus0.1)} \\
    \midrule
    \multirow{3}*{\shortstack{Tessar\\(1\texttimes\  res.)}} & \multirow{2}*{\lock} & \multirow{2}*{14.8} & \multirow{2}*{29.6} & \multirow{2}*{0.90} & \lock & 31.2 & 25.7 \\
    \arrayrulecolor{lightgray}\cmidrule(l{2pt}){6-8}\arrayrulecolor{black} &  &  &  &  & \unlock & 33.4 & 26.9 \\
    \arrayrulecolor{lightgray}\cmidrule(l{2pt}){2-8}\arrayrulecolor{black} & \unlock & 14.8 & 29.6 & 0.90 & \unlock & 33.6\color{Green}{(+0.2)} & 27.0\color{Green}{(+0.1)} \\
    \midrule
    \multirow{2}*{\shortstack{{\redstar}Tessar\\(2\texttimes\  res.)}} & \lock & 14.8 & 28.8 & 0.87 & \unlock & 31.3 & 24.4 \\
    \arrayrulecolor{lightgray}\cmidrule(l{2pt}){2-8}\arrayrulecolor{black} & \unlock & 24.7 & 27.9 & 0.85 & \unlock & 32.2\color{Green}{(+0.9)} & 26.2\color{Green}{(+1.8)} \\
    \bottomrule
\end{tabularx}

%% file: supplement.tex
This supplemental document provides additional information in support of the findings from the main manuscript.
Specifically, we describe the impact of the proposed ray-aiming approach in more detail, give further description of the wavelength selection for the proposed method, and discuss the assumptions made when deriving the optical image formation model.
Furthermore, we provide a tolerancing analysis, visualizations of our ablation experiments, additional comparisons to \citet{li2021end} and \citet{tseng2021differentiable}, and additional details on the experiments from the main manuscript.

\section{Code and Videos}
We provide access to our \href{https://github.com/princeton-computational-imaging/joint-lens-design}{\color{magenta}{code}}\footnote{\url{https://github.com/princeton-computational-imaging/joint-lens-design}} for simulating and optimizing compound refractive lenses in an end-to-end manner using exact differentiable ray tracing.

Additionally, our \href{https://light.princeton.edu/joint-lens-design}{\color{magenta}{project webpage}}\footnote{\url{https://light.princeton.edu/joint-lens-design}} includes videos that illustrate the joint optimization of the Doublet, Cooke, and Tessar lenses for object detection on the BDD100K dataset under regular (1\texttimes) resolution.
In particular, the videos illustrate how the selection of catalog glasses is handled through quantized continuous glass variables.
Whereas the Doublet lens constantly varies throughout the optimization process, the Cooke and Tessar lenses exhibit a different behavior in which the state of the lens changes sporadically and abruptly and then quickly stabilizes.
This behavior can be partly attributed to using the Adam optimizer~\cite{kingma2015adam} with high~$\beta_1$ and~$\beta_2$ values (0.9 and 0.999, respectively), where small perturbations can quickly add up due to slowly adapting learning rates.
We empirically found this behavior helpful in maximizing object detection performance, possibly due to beneficial regularization on the object detector.

\section{Ray Aiming}

\begin{figure}[t]
    \centering
    \includegraphics{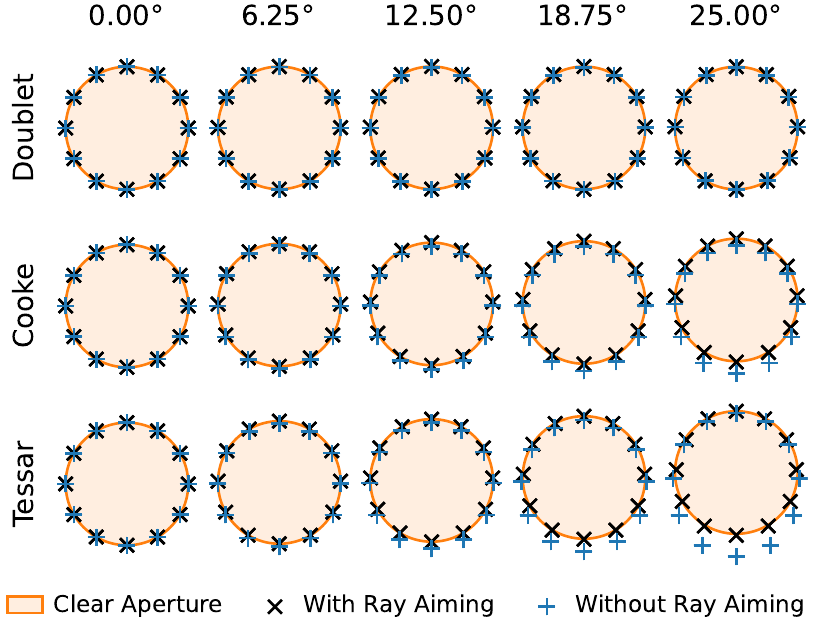}
    \caption{
        Ray-aiming error of rays aimed at the circular edge of the aperture stop with/without the ray-aiming correction step, for the f/2 Doublet, Cooke, and Tessar lenses used in our experiments and different field angles.
        While the ray-aiming correction step is not required for the Doublet, it prevents moderate errors with the Cooke lens and large errors with the Tessar lens.
    }
    \label{fig:ray_aiming_full}
\end{figure}

For any field angle within the full field of view, in the absence of optical vignetting, simulated rays incident upon a lens should precisely span the entire clear area of the aperture stop.
To fulfill this condition, it is common in conventional ray tracing to initialize the rays at the entrance pupil of the system, whose size and position are found through a paraxial ray-tracing operation.
Under strong pupil aberrations---that is, the aberrations between the entrance pupil and physical aperture stop---rays that are initialized naively at the entrance pupil may strongly deviate from their corresponding location on the aperture stop and, as such, skew the results of the ray-tracing operation.
In contrast to previous works that tackle the joint design of compound optics through ray tracing~\cite{li2021end,hale2021end}, we compensate for pupil aberrations with an accurate ray-aiming procedure, which consists in correcting the coordinates of the rays at the entrance pupil so that they adequately span the aperture stop.
Similar to~\citet{cote2021deep}, we assume an elliptic shape for the corrected entrance pupil; thus, we find approximations for the displacements of the top~$\Delta y_\mathrm{p, top}^h$, bottom~$\Delta y_\mathrm{p, bottom}^h$, and side~$\Delta x_\mathrm{p, side}^h$ of the pupil for each off-axis field value $h$.
Due to rotational symmetry, we have $\Delta x_\mathrm{p, right}^h = - \Delta x_\mathrm{p, left}^h = \Delta x_\mathrm{p, side}^h$.

As shown in \cref{fig:ray_aiming_full}, we find that the assumption of a linear relationship between the entrance pupil coordinates~$x_\mathrm{p}$, $y_\mathrm{p}$ and the aperture stop coordinates $x_\mathrm{s}$, $y_\mathrm{s}$ provides sufficient accuracy for the lenses used in our experiments, that is
\begin{align}
    \Delta x_\mathrm{s}
    &\approx
    \Delta x_\mathrm{p}
    \frac
    {\mathrm{d}x_\mathrm{s}}
    {\mathrm{d}x_\mathrm{p}}
    \,;
    \label{eq:ray_aiming_x}\\
    \Delta y_\mathrm{s}
    &\approx
    \Delta y_\mathrm{p}
    \frac
    {\mathrm{d}y_\mathrm{s}}
    {\mathrm{d}y_\mathrm{p}}
    \,.
    \label{eq:ray_aiming_y}
\end{align}
To compute the ray-aiming errors~$\Delta x_\mathrm{s, side}^h$, $\Delta y_\mathrm{s, top}^h$, and $\Delta y_\mathrm{s, bottom}^h$, we trace a sagittal ray and two meridional (upper and lower) rays for each field~$h$, respectively, then compare their coordinates at the aperture stop to the aperture stop diameter---computed by tracing an on-axis meridional ray.
The derivative terms are obtained through automatic differentiation.
Then, \cref{eq:ray_aiming_x} and \cref{eq:ray_aiming_y} are used to recover the field-wise entrance pupil displacements.

\section{Wavelength Selection}

\begin{figure}[t]
    \centering
    \includegraphics{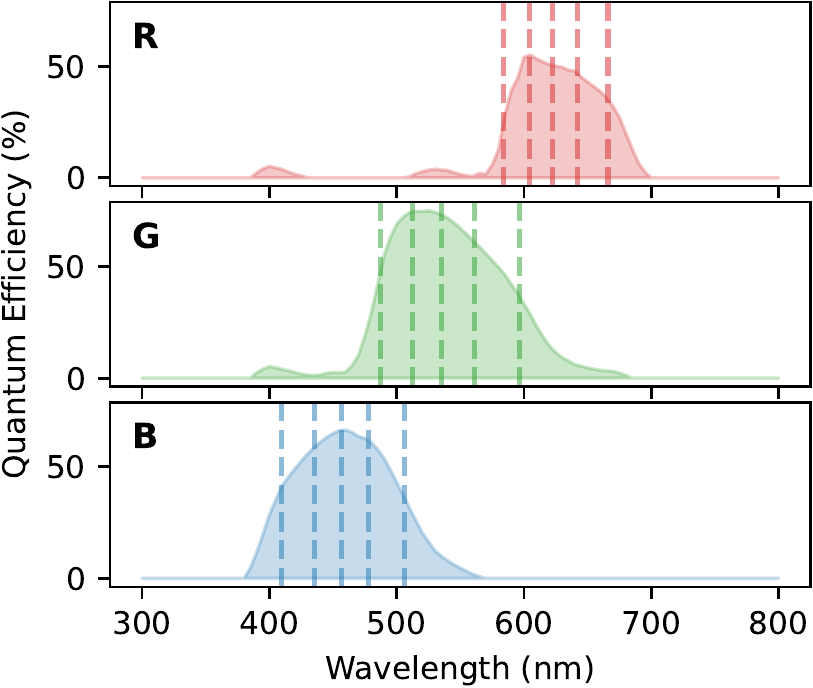}
    \caption{
        Selected wavelengths (indicated with vertical lines) for our experiments, which are based on the quantum efficiency spectrum of a typical sensor, here the Sony IMX172.
        }
    \label{fig:wavelength_selection}
\end{figure}

\begin{table}[t]
    \centering
    \begin{tabular}{ccc}
        \toprule
            R &     G &     B \\
        \midrule
        584.1 & 487.1 & 409.4 \\
        604.2 & 512.1 & 435.4 \\
        622.5 & 535.1 & 456.6 \\
        642.2 & 560.8 & 477.9 \\
        665.9 & 596.3 & 505.9 \\
        \bottomrule
    \end{tabular}
    \caption{
        Selected wavelengths (in \unit{\nm}) for each color channel~R,~G, and~B.
    }
    \label{tab:wavelength_selection}
\end{table}

In our experiments, we perform wavelength sampling that is representative of compound lenses.
To this end, we rely on the quantum efficiency spectrum~$Q(\lambda)$ of a typical sensor (here, the Sony IMX172), which is visualized in \cref{fig:wavelength_selection}.

For each of the R, G, and B color channels, we select 5~wavelengths by computing all the odd-numbered 10-quantiles of~$Q(\lambda)$.
The selected wavelengths are given in \cref{tab:wavelength_selection}.
While these wavelengths adequately represent the spectrum for our task, we note that the proposed method supports denser wavelength sampling without any changes, though at the cost of additional compute overhead.

\section{Image Formation Model}

Here we further discuss the assumptions made in the main paper and how they can impact our findings.

In our approach, we design all lenses for imaging at optical infinity.
This assumption---common and often safe in lens design and computational imaging---is adequate beyond the hyperfocal distance~$H = \nicefrac{f^2}{Nc}$, where $f$ is the focal length and $N$ is the f-number.
We can estimate the hyperfocal distance by setting an appropriate value for the tolerated circle of confusion.
As we consider that a circle of confusion smaller than the spot size diameter (i.e., twice the spot size radius) of a lens will have limited impact on optical performance, we set $c$ as the mean spot size diameter of the 2-, 3-, and 4-element baseline lenses, with
$f = \qty{17.2}{\mm}$ and $N = 2$ for all lenses, and estimate 0.9, 2.4, and \qty{5.0}{\m} for $H$, respectively.
For the envisioned object detection applications, most small objects that have to be located are typically found beyond this range, thus justifying this assumption.

\begin{figure}[t]
    \centering
    \includegraphics{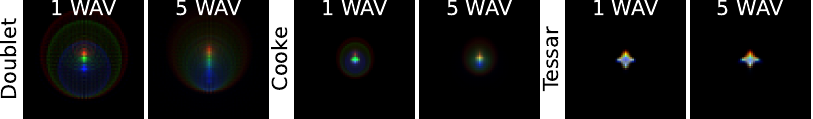}
    \caption{
        Multispectral sampling.
        The PSFs (shown at \ang{12.5} field angle) are \emph{more spread out} when chromatic aberrations are accurately captured with multispectral sampling (5 wavelengths).
        }
    \label{fig:spectrum}
\end{figure}

In the optical formation model used in this work, we implicitly consider the RGB values of an input image to be proportional to the luminance of the virtual scene.
An underlying assumption is that the spectra for each of the R, G, and B channels are uniform and do not depend on the scene content, which is not the case in practice.
Nonetheless, even though we contend with only three spectral bands (RGB), we accurately model chromatic aberrations with multispectral sampling.
We assume the worst-case scenario for broadband spectrum and \emph{overestimate} chromatic aberrations in the general case (see Fig.~\ref{fig:spectrum}).
As such, with real scenes rather than virtual ones, the actual chromatic aberrations would generally be smaller and object detection performance would presumably not be adversely impacted.

\section{Tolerancing Analysis}
We include a Monte-Carlo tolerancing analysis in Tab.~\ref{tab:tolerancing}, which supports that fabrication would marginally impact object detection performance while maintaining the margins that are reported in the main manuscript.
We note that object detection performance with our joint optimization method, compared to the alternative of fine-tuning the object detector but fixing the lens, does not suffer more from fabrication tolerances.

Precisely, before evaluating each image (n$=$\num{10000}), we apply random perturbations to each lens design by uniformly sampling across standard tolerances from Optimax and Schott (curvature: \qty{0.2}{\percent}; glass thickness: \qty{0.05}{\mm}; refractive index: \num{5e-4}; Abbe number: \qty{0.5}{\percent}).
We note that our designs are sensitive to changes in glass thickness due to their small total track length, so we employ only the second-most economical option (\qty{0.05}{\mm} instead of \qty{0.15}{\mm}).
In this analysis, we neglect airspace tolerances and use the paraxial image solve as we consider that the focus could be adjusted during fabrication.

Incidentally, we note that hypothetical discrepancies between the designed and manufactured lens could be accounted for by fine-tuning the downstream model using the measured optical performance metrics (PSF, distortion, etc.), as is common in joint design methodology~\cite{tseng2021differentiable}.

\begin{table}[t]
    \centering
    \small
    \renewcommand*{\arraystretch}{1}
    \setlength{\tabcolsep}{0em}
    \begin{tabularx}{\linewidth}{lXcXcXcXc}
        \toprule
        Optics && Baseline w/ && Proposed w/ && Margin && Margin w/ \\
         && tolerancing && tolerancing &&  && tolerancing \\
        \midrule
        Doublet (1\texttimes\@ res.) && 30.3 \color{gray} (-0.0) && 32.0 \color{gray} (-0.0) && \color{Green}+1.7 && \color{Green}+1.7 \\
        Cooke (1\texttimes\@ res.) && 33.0 \color{gray} (-0.0) && 33.3 \color{gray} (-0.0) && \color{Green}+0.3 && \color{Green}+0.3 \\
        Tessar (1\texttimes\@ res.) && 33.4 \color{gray} (-0.0) && 33.6 \color{gray} (-0.0) && \color{Green}+0.2 && \color{Green}+0.2 \\ 
        Doublet (2\texttimes\@ res.) && 25.0 \color{gray} (-0.0) && 28.1 \color{gray} (-0.0) && \color{Green}+3.1 && \color{Green}+3.1 \\
        Cooke (2\texttimes\@ res.) && 31.5 \color{gray} (-0.0) && 31.7 \color{gray} (-0.0) && \color{Green}+0.2 && \color{Green}+0.2 \\
        Tessar (2\texttimes\@ res.) && 31.1 \color{red} (-0.2) && 32.1 \color{red} (-0.1) && \color{Green}+0.9 && \color{Green}+1.0 \\ 
        \bottomrule
    \end{tabularx}
    \caption{
        Monte-Carlo tolerancing analysis (n$=$\num{10000}).
        We report the change in average AP on the BDD100K dataset w.r.t.\@ Tab.\@ 2 of the main paper when including tolerancing.
        We note that the Tessar lens is more sensible to fabrication tolerances due to having more lens elements.
        }
    \label{tab:tolerancing}
\end{table}

\section{Qualitative Ablation Experiments}

\begin{figure}[t]
    \centering
    \includegraphics{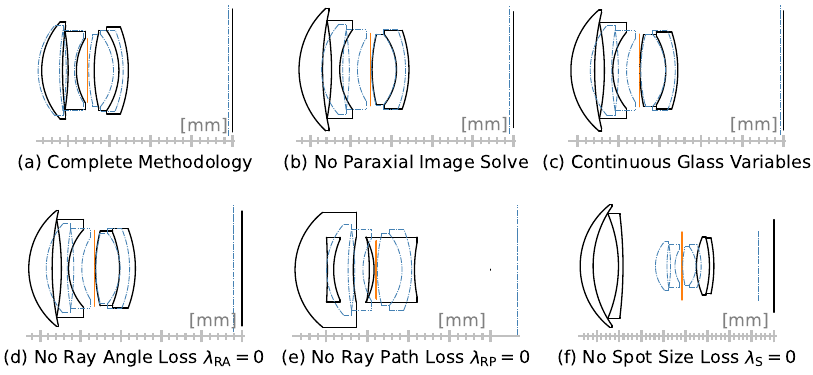}
    \caption{
        Qualitative ablation experiments.
        Illustrated are the lens layouts after the joint optimization of the Tessar lens on the BDD100K dataset under 2\texttimes\ simulated resolution.
        In this experimental setting, our complete methodology (a) favors a lens that resembles the baseline lens and starting point (shown in blue with the aperture stop as the reference plane).
        In~(b), (c), and~(d), in addition to vignetted rays (not shown), the lens deviates further from the starting point and ends up with strong pupil aberrations that cannot be handled by a single ray-aiming correction step.
        In~(e), overlapping lens elements result from the removal of the ray path loss.
        In~(f), without the spot size loss, only the noisy object detection loss drives the optimization process; as a result, the lens diverges significantly from the starting point and ends up with a mean spot size 9.6 times the one of the baseline lens.
        }
    \label{fig:ablation_layouts}
\end{figure}

In \cref{fig:ablation_layouts}, we report the final 2D lens layouts that accompany each of the ablation experiments from the main document, for the joint design of the Tessar lens on 2\texttimes\ simulated resolution.
For this experimental setting, \emph{removing any component of the proposed method harms the stability} of the optimization process and results in a poorly behaved lens.

\section{Comparison to \citet{li2021end}}

In this section, we present additional comparisons to the ray-tracing approach proposed by \citet{li2021end}.
In \cref{tab:comparison_li}, we report Tessar lens experiments by making two changes to our ray-tracing algorithm: as in~\cite{li2021end}, we fill the entrance pupil with a square grid of rays (such that the corners of the square grid hit the circular edge of the aperture stop), and ignore accurate ray aiming.
We train the Tessar lens and object detector jointly using this modified ray-tracing (MRT) algorithm, then evaluate the trained model using both the MRT algorithm and our complete ray-tracing (CRT) algorithm.
Under both 1\texttimes\ and 2\texttimes\ simulated resolution, the MRT algorithm leads to an underestimated spot size (\qty{14.2}{\um} and \qty{21.0}{\um} instead of \qty{16.0}{\um} and \qty{26.5}{\um} on 1\texttimes\ and 2\texttimes\ simulated resolution, respectively).
Likewise, the average precision (AP) is overestimated when using the MRT instead of the CRT (33.4 and 28.3 instead of 33.2 and 27.5, respectively).
This validates the proposed method as a more accurate ray-tracing algorithm to investigate task-driven optical design.

\begin{table}[tb]
    \centering
    \footnotesize
    \setlength{\tabcolsep}{0em}
    \begin{tabularx}{\linewidth}{cXcXcXcXcXc}
        \toprule
        &&&& \multicolumn{3}{c}{Eval. with MRT} && \multicolumn{3}{c}{Eval. with CRT} \\
        \arrayrulecolor{lightgray}\cmidrule{5-7}\cmidrule{9-11}\arrayrulecolor{black} Setting && Optics && Spot (\si{\um})$_\downarrow$ && AP$_\uparrow$ && Spot (\si{\um})$_\downarrow$ && AP$_\uparrow$ \\
        \midrule
        MRT~\cite{li2021end} && \multirow{2}*{Tessar (1\texttimes\ res.)} && 14.2 && 33.4 && 16.0 && 33.2 \\
        \textbf{Ours} &&&&&&&& 14.8 && \textbf{33.6} \\
        \midrule
        MRT~\cite{li2021end} && \multirow{2}*{Tessar (2\texttimes\ res.)} && 21.0 && 28.3 && 26.5 && 27.5 \\
        \textbf{Ours} &&&&&&&& 24.7 && \textbf{32.2} \\
        \bottomrule
    \end{tabularx}
    \caption{
        Comparison on the joint optimization of the Tessar lens with a modified ray-tracing (MRT) algorithm that follows the methodology in~\citet{li2021end}.
        We report the AP on BDD100K, where aberrations are modeled using either the MRT algorithm or our complete ray-tracing (CRT) algorithm.
        We also report the mean spot size evaluated using both ray-tracing algorithms.
        }
    \label{tab:comparison_li}
\end{table}

\begin{figure*}[tb]
    \centering
    \includegraphics{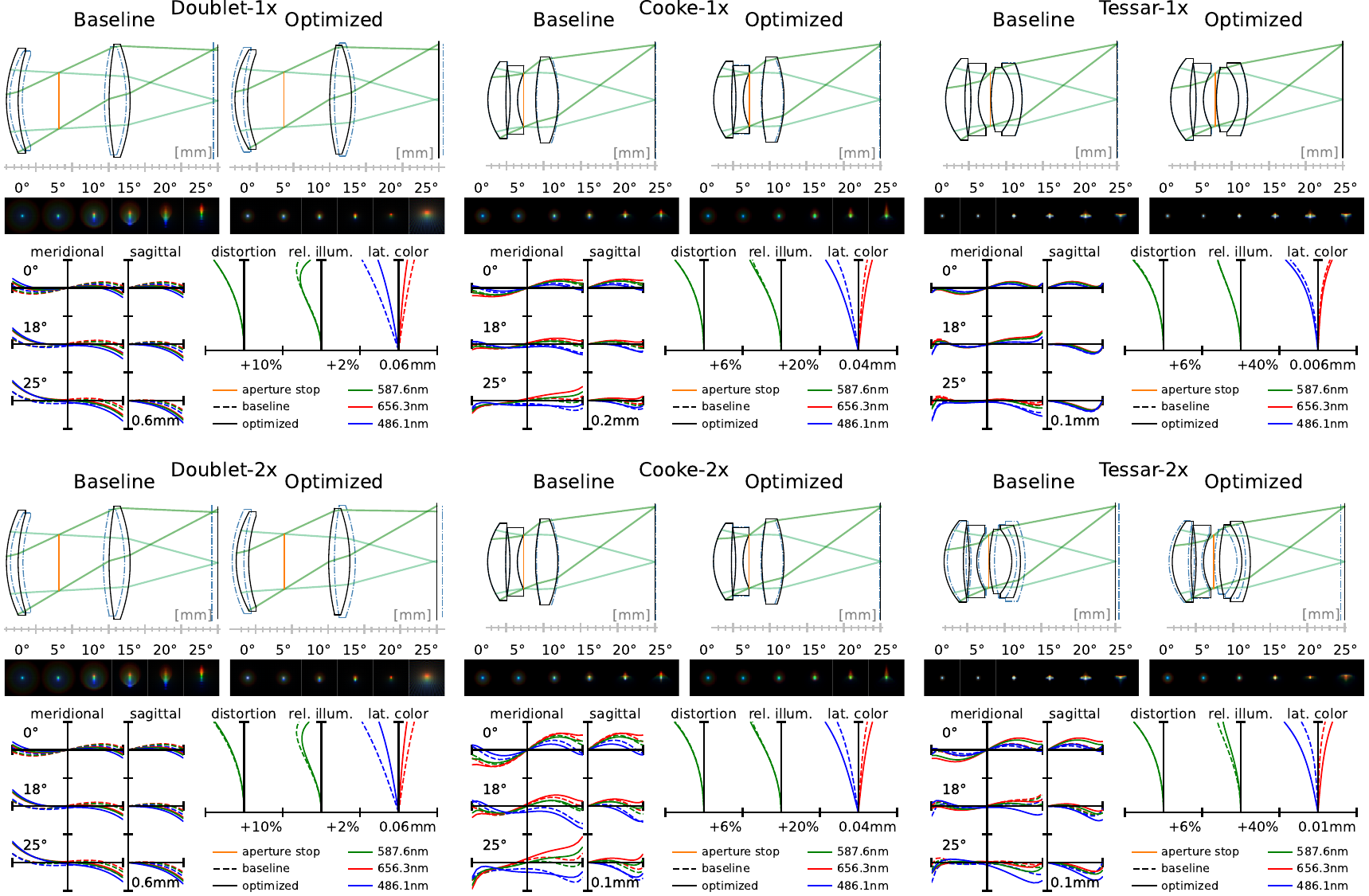}
    \caption{
        Baseline and optimized lenses.
        From top to bottom, we show 1)~the lens designs (dashed lines represent the baseline/optimized counterpart); 2)~PSFs for different fields; and 3)~aberration charts (left: ray fan plots; right: field curves).
    }
    \label{fig:qualitative_all}
\end{figure*}

\begin{table}[tb]
    \centering
    \small
    \renewcommand*{\arraystretch}{0.875}
    \setlength{\tabcolsep}{0.25em}
    \input{proxy_limits.tex}
    \caption{
        Predefined boundaries for each of the 22~normalized Tessar lens variables, which we use to replicate the proxy model approach of \citet{tseng2021differentiable}.
        As in~\cite{tseng2021differentiable}, the boundaries have two purposes: to sample the lens variables used to train the proxy model, and to limit the allowed range during joint optimization experiments.
        Note that there is no curvature for the flat aperture stop (denoted *) nor for the last optical surface, which is computed using a paraxial ray-tracing operation to enforce the desired focal length.
        }
    \label{tab:proxy_limits}
\end{table}

\begin{table*}[tb]
    \centering
    \footnotesize
    \renewcommand*{\arraystretch}{0.875}
    \setlength{\tabcolsep}{.5em}
    \input{lens_parameters.tex}
    \caption{
        Complete list of lens parameters for all experiments: radii~$\nicefrac{1}{c}$, spacings~$s$, and glass materials along with the refractive index~$n_\mathrm{d}$ and Abbe number~$v_\mathrm{d}$.
        The aperture stop surface is denoted with~*.
    }
    \label{tab:lens_parameters}
\end{table*}

\section{Proxy Model}

In this section, we provide additional details on our comparison with the proxy model of \citet{tseng2021differentiable}, closely adapted here for fair comparison.

We first generate 10k variations of the baseline Tessar lens by uniformly sampling each of the 22~lens variables between predefined boundaries as given in~\cref{tab:proxy_limits}.
The variable boundaries are centered on the lens parameters of the baseline Tessar lens.
The allowed range for each lens variable is set to 0.4 times the standard deviation of each variable group: 6~normalized curvatures~$c'$, 8~normalized spacings~$s'$, and 8~normalized glass variables~$g$.
In joint optimization experiments, these variable boundaries are also used to clip each lens variable after each optimization step.
In contrast to the use of a proxy model, we note that our ray-tracing approach does not require predefined boundaries: the ray-tracing algorithm works for all of the solution space (as long as ray aiming remains accurate), and manufacturing constraints are handled with carefully designed losses instead of limiting the solution space within a predefined region.

Quantized continuous glass variables do not synergize well with a proxy model, so we use standard continuous relaxations instead.
However, we do use the paraxial image solve as in other experiments.

\paragraph{Model Architecture and Training}
As in~\citet{tseng2021differentiable}, our proxy model consists of a multilayer perceptron (MLP) followed by a convolutional decoder.
The MLP takes as inputs the 22 lens variables as well as the field value, and is composed of two hidden layers with 128 units and an output layer with $32\cdot32\cdot3 + 2 = 3074$ units.
The last 2~units are used for the relative illumination factor and distortion shift.
The rest are reshaped (\numproduct{32x32x3}), then fed into the decoder with the following architecture:

\begin{itemize}
    \item two \numproduct{3x3} convolutional layers with 64 output channels (output size is \numproduct{32x32x64});
    \item transposed convolutional layer for 2\texttimes\ upsampling;
    \item two \numproduct{3x3} convolutional layers with 64 output channels (output size is \numproduct{64x64x64});
    \item transposed convolutional layer for 1\texttimes\ upsampling to closely follow Tseng despite a smaller PSF size (output size is \numproduct{65x65x64});
    \item \numproduct{3x3} convolutional layer with 3 output channels (output size is \numproduct{65x65x3}).
\end{itemize}

We obtain the PSFs by normalizing the outputs using a softmax operation so that their channel-wise area is~1, as is the case for the ground truth PSFs.
The proxy model is trained on 10~epochs using the Adam optimizer~\cite{kingma2015adam} with a learning rate of \qty{0.001} and a batch size of~10.

\section{Additional Results}

\Cref{fig:qualitative_all} provides detailed aberration charts and lens layouts for the Doublet, Cooke, and Tessar lenses reported in the main paper, optimized either for spot size (baseline) or object detection with 1\texttimes\ or 2\texttimes\ simulated resolution.
\Cref{tab:lens_parameters} lists the corresponding lens parameters for each experimental setting.

%% file: proxy_limits.tex
\begin{tabularx}{\linewidth}{cXccXccXccXcc}
    \toprule
     &  & \multicolumn{2}{c}{$c'$} &  & \multicolumn{2}{c}{$s'$} &  & \multicolumn{2}{c}{$g_1$} &  & \multicolumn{2}{c}{$g_2$} \\
     &  & min & max &  & min & max &  & min & max &  & min & max \\
    \midrule
    1 &  & 1.72 & 2.37 &  & 0.140 & 0.164 &  & -1.17 & -0.71 &  & -1.11 & -0.65 \\
    2 &  & 0.28 & 0.93 &  & 0.035 & 0.059 &  &  &  &  &  &  \\
    3 &  & -0.88 & -0.23 &  & 0.046 & 0.071 &  & -1.14 & -0.68 &  & 0.93 & 1.39 \\
    4 &  & 1.96 & 2.62 &  & 0.081 & 0.105 &  &  &  &  &  &  \\
    5* &  &  &  &  & -0.004 & 0.021 &  &  &  &  &  &  \\
    6 &  & 0.66 & 1.31 &  & 0.163 & 0.187 &  & -1.17 & -0.71 &  & -1.11 & -0.65 \\
    7 &  & -2.81 & -2.16 &  & 0.056 & 0.081 &  & -3.64 & -3.18 &  & -0.93 & -0.47 \\
    8 &  &  &  &  & -0.028 & -0.004 &  &  &  &  &  &  \\
    \bottomrule
\end{tabularx}

%% file: lens_parameters.tex
\begin{tabularx}{\linewidth}{cXcccccXcccccXccccc}
    \toprule
     &  & 1/$c$ & $s$ & Glass & $n_\mathrm{d}$ & $v_\mathrm{d}$ &  & 1/$c$ & $s$ & Glass & $n_\mathrm{d}$ & $v_\mathrm{d}$ &  & 1/$c$ & $s$ & Glass & $n_\mathrm{d}$ & $v_\mathrm{d}$ \\
     &  & mm & mm &  &  &  &  & mm & mm &  &  &  &  & mm & mm &  &  &  \\
    \midrule
     &  & \multicolumn{5}{c}{\textbf{Doublet (Baseline)}} &  & \multicolumn{5}{c}{\textbf{Doublet (Optimized, 1\texttimes\ res.)}} &  & \multicolumn{5}{c}{\textbf{Doublet (Optimized, 2\texttimes\ res.)}} \\
    \midrule
    1 &  & 16.71 & 1.61 & S-LAL12 & 1.678 & 55.3 &  & 13.14 & 1.73 & S-TIM1 & 1.626 & 35.7 &  & 14.01 & 1.72 & S-TIM1 & 1.626 & 35.7 \\
    2 &  & 22.92 & 5.60 &  &  &  &  & 17.78 & 5.03 &  &  &  &  & 19.83 & 4.94 &  &  &  \\
    3* &  & inf & 6.90 &  &  &  &  & inf & 6.27 &  &  &  &  & inf & 6.12 &  &  &  \\
    4 &  & 44.34 & 2.89 & S-LAH92 & 1.892 & 37.1 &  & 32.88 & 2.99 & S-LAH96 & 1.764 & 48.5 &  & 33.02 & 2.92 & S-LAH96 & 1.764 & 48.5 \\
    5 &  & -22.87 & 12.03 &  &  &  &  & -21.76 & 12.00 &  &  &  &  & -22.17 & 12.00 &  &  &  \\
    \midrule
     &  & \multicolumn{5}{c}{\textbf{Cooke (Baseline)}} &  & \multicolumn{5}{c}{\textbf{Cooke (Optimized, 1\texttimes\ res.)}} &  & \multicolumn{5}{c}{\textbf{Cooke (Optimized, 2\texttimes\ res.)}} \\
    \midrule
    1 &  & 9.10 & 2.44 & S-LAH96 & 1.764 & 48.5 &  & 8.98 & 2.44 & S-LAH96 & 1.764 & 48.5 &  & 8.98 & 2.44 & S-LAH96 & 1.764 & 48.5 \\
    2 &  & 67.86 & 0.57 &  &  &  &  & 58.92 & 0.57 &  &  &  &  & 58.52 & 0.57 &  &  &  \\
    3 &  & -26.08 & 1.00 & S-TIM1 & 1.626 & 35.7 &  & -27.40 & 1.00 & S-TIM1 & 1.626 & 35.7 &  & -27.33 & 1.00 & S-TIM1 & 1.626 & 35.7 \\
    4 &  & 8.35 & 0.84 &  &  &  &  & 8.40 & 0.83 &  &  &  &  & 8.40 & 0.83 &  &  &  \\
    5* &  & inf & 1.60 &  &  &  &  & inf & 1.71 &  &  &  &  & inf & 1.73 &  &  &  \\
    6 &  & 25.01 & 3.00 & S-LAH96 & 1.764 & 48.5 &  & 29.63 & 3.00 & S-LAH92 & 1.892 & 37.1 &  & 29.35 & 3.00 & S-LAH92 & 1.892 & 37.1 \\
    7 &  & -15.20 & 13.06 &  &  &  &  & -17.97 & 13.06 &  &  &  &  & -18.02 & 13.09 &  &  &  \\
    \midrule
     &  & \multicolumn{5}{c}{\textbf{Tessar (Baseline)}} &  & \multicolumn{5}{c}{\textbf{Tessar (Optimized, 1\texttimes\ res.)}} &  & \multicolumn{5}{c}{\textbf{Tessar (Optimized, 2\texttimes\ res.)}} \\
    \midrule
    1 &  & 8.39 & 2.61 & S-LAH96 & 1.764 & 48.5 &  & 8.40 & 2.61 & S-LAH96 & 1.764 & 48.5 &  & 8.89 & 2.53 & S-LAH96 & 1.764 & 48.5 \\
    2 &  & 28.27 & 0.81 &  &  &  &  & 28.14 & 0.81 &  &  &  &  & 39.54 & 0.68 &  &  &  \\
    3 &  & -30.99 & 1.00 & S-TIM1 & 1.626 & 35.7 &  & -31.12 & 1.00 & S-TIM1 & 1.626 & 35.7 &  & -30.34 & 1.00 & S-TIL25 & 1.581 & 40.7 \\
    4 &  & 7.49 & 1.60 &  &  &  &  & 7.47 & 1.62 &  &  &  &  & 7.69 & 1.36 &  &  &  \\
    5* &  & inf & 0.14 &  &  &  &  & inf & 0.14 &  &  &  &  & inf & 0.80 &  &  &  \\
    6 &  & 17.43 & 3.00 & S-LAH96 & 1.764 & 48.5 &  & 17.33 & 3.00 & S-LAH96 & 1.764 & 48.5 &  & 19.53 & 3.00 & S-LAH96 & 1.764 & 48.5 \\
    7 &  & -6.90 & 1.17 & S-LAH88 & 1.916 & 31.6 &  & -6.95 & 1.22 & S-LAH88 & 1.916 & 31.6 &  & -8.72 & 1.15 & S-LAH88 & 1.916 & 31.6 \\
    8 &  & -13.00 & 12.84 &  &  &  &  & -13.02 & 12.86 &  &  &  &  & -14.85 & 12.68 &  &  &  \\
    \bottomrule
    \end{tabularx}

%% file: main.bbl
\begin{thebibliography}{45}
\providecommand{\natexlab}[1]{#1}
\providecommand{\url}[1]{\texttt{#1}}
\expandafter\ifx\csname urlstyle\endcsname\relax
  \providecommand{\doi}[1]{doi: #1}\else
  \providecommand{\doi}{doi: \begingroup \urlstyle{rm}\Url}\fi

\bibitem[Abadi et~al.(2016)Abadi, Agarwal, Barham, Brevdo, Chen, Citro,
  Corrado, Davis, Dean, Devin, Ghemawat, Goodfellow, Harp, Irving, Isard, Jia,
  Jozefowicz, Kaiser, Kudlur, Levenberg, Man\'{e}, Monga, Moore, Murray, Olah,
  Schuster, Shlens, Steiner, Sutskever, Talwar, Tucker, Vanhoucke, Vasudevan,
  Vi\'{e}gas, Vinyals, Warden, Wattenberg, Wicke, Yu, and
  Zheng]{i2016tensorflow}
M.~Abadi, A.~Agarwal, P.~Barham, E.~Brevdo, Z.~Chen, C.~Citro, G.~S. Corrado,
  A.~Davis, J.~Dean, M.~Devin, S.~Ghemawat, I.~Goodfellow, A.~Harp, G.~Irving,
  M.~Isard, Y.~Jia, R.~Jozefowicz, L.~Kaiser, M.~Kudlur, J.~Levenberg,
  D.~Man\'{e}, R.~Monga, S.~Moore, D.~Murray, C.~Olah, M.~Schuster, J.~Shlens,
  B.~Steiner, I.~Sutskever, K.~Talwar, P.~Tucker, V.~Vanhoucke, V.~Vasudevan,
  F.~Vi\'{e}gas, O.~Vinyals, P.~Warden, M.~Wattenberg, M.~Wicke, Y.~Yu, and
  X.~Zheng.
\newblock {TensorFlow}: Large-scale machine learning on heterogeneous systems.
\newblock \emph{arXiv:1603.04467}, 2016.

\bibitem[Bentley and Olson(2012)]{bentley2012field}
J.~Bentley and C.~Olson.
\newblock Field guide to lens design.
\newblock Society of Photo-Optical Instrumentation Engineers (SPIE), 2012.

\bibitem[Chang and Wetzstein(2019)]{chang2019deep}
J.~Chang and G.~Wetzstein.
\newblock Deep {Optics} for {Monocular} {Depth} {Estimation} and {3D} {Object}
  {Detection}.
\newblock In \emph{Proceedings of {IEEE} {International} {Conference} on
  {Computer} {Vision}}, pages 10192--10201. IEEE, 2019.
\newblock \doi{10.1109/ICCV.2019.01029}.

\bibitem[Chang et~al.(2018)Chang, Sitzmann, Dun, Heidrich, and
  Wetzstein]{chang2018hybrid}
J.~Chang, V.~Sitzmann, X.~Dun, W.~Heidrich, and G.~Wetzstein.
\newblock Hybrid optical-electronic convolutional neural networks with
  optimized diffractive optics for image classification.
\newblock \emph{Scientific reports}, 8\penalty0 (1):\penalty0 1--10, 2018.

\bibitem[C\^{o}t\'{e} et~al.(2021{\natexlab{a}})C\^{o}t\'{e}, Lalonde, and
  Thibault]{cote2021deep}
G.~C\^{o}t\'{e}, J.-F. Lalonde, and S.~Thibault.
\newblock Deep learning-enabled framework for automatic lens design starting
  point generation.
\newblock \emph{Opt. Express}, 29\penalty0 (3):\penalty0 3841--3854, Feb
  2021{\natexlab{a}}.
\newblock \doi{10.1364/OE.401590}.

\bibitem[C\^{o}t\'{e} et~al.(2021{\natexlab{b}})C\^{o}t\'{e}, Lalonde, and
  Thibault]{cote2021on}
G.~C\^{o}t\'{e}, J.-F. Lalonde, and S.~Thibault.
\newblock {On the use of deep learning for lens design}.
\newblock \emph{Proc. SPIE}, 12078:\penalty0 230 -- 236, 2021{\natexlab{b}}.
\newblock \doi{10.1117/12.2603656}.

\bibitem[C{\^o}t{\'e} et~al.(2022)C{\^o}t{\'e}, Zhang, Menke, Lalonde, and
  Thibault]{cote2022inferring}
G.~C{\^o}t{\'e}, Y.~Zhang, C.~Menke, J.-F. Lalonde, and S.~Thibault.
\newblock Inferring the solution space of microscope objective lenses using
  deep learning.
\newblock \emph{Optics Express}, 30\penalty0 (5):\penalty0 6531--6545, Feb.
  2022.
\newblock ISSN 1094-4087.
\newblock \doi{10.1364/OE.451327}.

\bibitem[Côté(2021)]{cote2021lensnet}
G.~Côté.
\newblock {LensNet}: lens design starting point generator, 2021.
\newblock \url{https://lvsn.github.io/lensnet}.

\bibitem[Girard(1958)]{girard1958excerpt}
A.~Girard.
\newblock Excerpt from {Revue} d'optique théorique et instrumentale.
\newblock \emph{Rev. Opt}, 37:\penalty0 225--241, 1958.

\bibitem[Haim et~al.(2018)Haim, Elmalem, Giryes, Bronstein, and
  Marom]{haim2018depth}
H.~Haim, S.~Elmalem, R.~Giryes, A.~M. Bronstein, and E.~Marom.
\newblock Depth {Estimation} {From} a {Single} {Image} {Using} {Deep} {Learned}
  {Phase} {Coded} {Mask}.
\newblock \emph{IEEE Transactions on Computational Imaging}, 4\penalty0
  (3):\penalty0 298--310, Sept. 2018.
\newblock ISSN 2333-9403.
\newblock \doi{10.1109/TCI.2018.2849326}.

\bibitem[Hal{\'e} et~al.(2021)Hal{\'e}, Trouv{\'e}-Peloux, and
  Volatier]{hale2021end}
A.~Hal{\'e}, P.~Trouv{\'e}-Peloux, and J.-B. Volatier.
\newblock End-to-end sensor and neural network design using differential ray
  tracing.
\newblock \emph{Optics express}, 29\penalty0 (21):\penalty0 34748--34761, 2021.

\bibitem[Hanika and Dachsbacher(2014)]{hanika2014efficient}
J.~Hanika and C.~Dachsbacher.
\newblock Efficient monte carlo rendering with realistic lenses.
\newblock In \emph{Computer Graphics Forum}, volume~33, pages 323--332. Wiley
  Online Library, 2014.

\bibitem[He et~al.(2016)He, Zhang, Ren, and Sun]{he2016deep}
K.~He, X.~Zhang, S.~Ren, and J.~Sun.
\newblock Deep residual learning for image recognition.
\newblock In \emph{Proceedings of the IEEE conference on computer vision and
  pattern recognition}, pages 770--778, 2016.

\bibitem[Hirsch et~al.(2010)Hirsch, Sra, Sch{\"o}lkopf, and
  Harmeling]{hirsch2010efficient}
M.~Hirsch, S.~Sra, B.~Sch{\"o}lkopf, and S.~Harmeling.
\newblock Efficient filter flow for space-variant multiframe blind
  deconvolution.
\newblock In \emph{2010 IEEE Computer Society Conference on Computer Vision and
  Pattern Recognition}, pages 607--614. IEEE, 2010.

\bibitem[Kingma and Ba(2015)]{kingma2015adam}
D.~P. Kingma and J.~Ba.
\newblock Adam: {A} {Method} for {Stochastic} {Optimization}.
\newblock In \emph{Proccedings of the 3rd {International} {Conference} on
  {Learning} {Representations}}, 2015.

\bibitem[Kolb et~al.(1995)Kolb, Mitchell, and Hanrahan]{kolb1995realistic}
C.~Kolb, D.~Mitchell, and P.~Hanrahan.
\newblock A realistic camera model for computer graphics.
\newblock In \emph{Proceedings of the 22nd annual conference on computer
  graphics and interactive techniques}, pages 317--324, 1995.

\bibitem[Li et~al.(2021)Li, Hou, Wang, Tan, Liu, and Zhang]{li2021end}
Z.~Li, Q.~Hou, Z.~Wang, F.~Tan, J.~Liu, and W.~Zhang.
\newblock End-to-end learned single lens design using fast differentiable ray
  tracing.
\newblock \emph{Optics Letters}, 46\penalty0 (21):\penalty0 5453--5456, 2021.

\bibitem[Lin et~al.(2017)Lin, Goyal, Girshick, He, and
  Doll{\'a}r]{lin2017focal}
T.-Y. Lin, P.~Goyal, R.~Girshick, K.~He, and P.~Doll{\'a}r.
\newblock Focal loss for dense object detection.
\newblock In \emph{Proceedings of the IEEE international conference on computer
  vision}, pages 2980--2988, 2017.

\bibitem[Maeda et~al.(2005)Maeda, Catrysse, and Wandell]{maeda2005integrating}
P.~Y. Maeda, P.~B. Catrysse, and B.~A. Wandell.
\newblock Integrating lens design with digital camera simulation.
\newblock In \emph{Digital {Photography}}, volume 5678, pages 48--58. SPIE,
  Feb. 2005.
\newblock \doi{10.1117/12.588153}.

\bibitem[Metzler et~al.(2020)Metzler, Ikoma, Peng, and
  Wetzstein]{metzler2020deep}
C.~A. Metzler, H.~Ikoma, Y.~Peng, and G.~Wetzstein.
\newblock Deep {Optics} for {Single}-{Shot} {High}-{Dynamic}-{Range} {Imaging}.
\newblock In \emph{Proceedings of {IEEE} {Conference} on {Computer} {Vision}
  and {Pattern} {Recognition}}, pages 1375--1385. IEEE, 2020.

\bibitem[{Ohara Corporation}(2019)]{corporation2019optical2}
{Ohara Corporation}.
\newblock Optical {Glass} {Catalog}, 2019.

\bibitem[Paszke et~al.(2019)Paszke, Gross, Massa, Lerer, Bradbury, Chanan,
  Killeen, Lin, Gimelshein, Antiga, Desmaison, Kopf, Yang, DeVito, Raison,
  Tejani, Chilamkurthy, Steiner, Fang, Bai, and Chintala]{paszke2019pytorch}
A.~Paszke, S.~Gross, F.~Massa, A.~Lerer, J.~Bradbury, G.~Chanan, T.~Killeen,
  Z.~Lin, N.~Gimelshein, L.~Antiga, A.~Desmaison, A.~Kopf, E.~Yang, Z.~DeVito,
  M.~Raison, A.~Tejani, S.~Chilamkurthy, B.~Steiner, L.~Fang, J.~Bai, and
  S.~Chintala.
\newblock Pytorch: An imperative style, high-performance deep learning library.
\newblock In \emph{Advances in neural information processing systems},
  volume~32, pages 8024--8035, 2019.

\bibitem[Peng et~al.(2019)Peng, Sun, Dun, Wetzstein, Heidrich, and
  Heide]{peng2019learned}
Y.~Peng, Q.~Sun, X.~Dun, G.~Wetzstein, W.~Heidrich, and F.~Heide.
\newblock Learned large field-of-view imaging with thin-plate optics.
\newblock \emph{ACM Trans. Graph.}, 38\penalty0 (6):\penalty0 219--1, 2019.

\bibitem[Rimmer(1986)]{rimmer1986relative}
M.~P. Rimmer.
\newblock {Relative Illumination Calculations}.
\newblock In R.~E. Fischer and P.~J. Rogers, editors, \emph{Optical System
  Design, Analysis, Production for Advanced Technology Systems}, volume 0655,
  pages 99 -- 104. International Society for Optics and Photonics, SPIE, 1986.
\newblock \doi{10.1117/12.938414}.

\bibitem[{Schott Corporation}(2019)]{corporation2019optical}
{Schott Corporation}.
\newblock Optical {Glass} {Catalog}, 2019.

\bibitem[Schuhmann(2019)]{schuhmann2019description}
R.~Schuhmann.
\newblock Description of aspheric surfaces.
\newblock \emph{Advanced Optical Technologies}, 8\penalty0 (3-4):\penalty0
  267--278, 2019.

\bibitem[Sitzmann et~al.(2018)Sitzmann, Diamond, Peng, Dun, Boyd, Heidrich,
  Heide, and Wetzstein]{sitzmann2018end}
V.~Sitzmann, S.~Diamond, Y.~Peng, X.~Dun, S.~Boyd, W.~Heidrich, F.~Heide, and
  G.~Wetzstein.
\newblock End-to-end optimization of optics and image processing for achromatic
  extended depth of field and super-resolution imaging.
\newblock \emph{ACM Transactions on Graphics}, 37\penalty0 (4):\penalty0 1--13,
  Aug. 2018.
\newblock ISSN 0730-0301, 1557-7368.
\newblock \doi{10.1145/3197517.3201333}.

\bibitem[Smith(2004)]{smith2004modern}
W.~J. Smith.
\newblock \emph{Modern {Lens} {Design}}.
\newblock McGraw Hill Professional, Nov. 2004.
\newblock ISBN 978-0-07-177726-1.

\bibitem[Steinert et~al.(2011)Steinert, Dammertz, Hanika, and
  Lensch]{steinert2011general}
B.~Steinert, H.~Dammertz, J.~Hanika, and H.~P. Lensch.
\newblock General spectral camera lens simulation.
\newblock In \emph{Computer Graphics Forum}, volume~30, pages 1643--1654. Wiley
  Online Library, 2011.

\bibitem[Sturlesi and O'Shea(1991)]{sturlesi1991future}
D.~Sturlesi and D.~C. O'Shea.
\newblock Future of global optimization in optical design.
\newblock In \emph{1990 {Intl} {Lens} {Design} {Conf}}, volume 1354, pages
  54--69. International Society for Optics and Photonics, Jan. 1991.
\newblock \doi{10.1117/12.47876}.

\bibitem[Sun et~al.(2020)Sun, Tseng, Fu, Heidrich, and Heide]{sun2020learning}
Q.~Sun, E.~Tseng, Q.~Fu, W.~Heidrich, and F.~Heide.
\newblock Learning {Rank}-1 {Diffractive} {Optics} for {Single}-{Shot} {High}
  {Dynamic} {Range} {Imaging}.
\newblock In \emph{Proceedings of {IEEE} {Conference} on {Computer} {Vision}
  and {Pattern} {Recognition}}, pages 1386--1396. IEEE, 2020.

\bibitem[Sun et~al.(2021)Sun, Wang, Fu, Dun, and Heidrich]{sun2021end}
Q.~Sun, C.~Wang, Q.~Fu, X.~Dun, and W.~Heidrich.
\newblock End-to-end complex lens design with differentiate ray tracing.
\newblock \emph{ACM Trans. Graph.}, 40\penalty0 (4):\penalty0 1--13, jul 2021.
\newblock ISSN 0730-0301.
\newblock \doi{10.1145/3450626.3459674}.

\bibitem[{Synopsys}(2018)]{synopsys2018code}
{Synopsys}.
\newblock Code {V} 11.2 {Documentation} {Library}.
\newblock Technical report, Synopsys, 2018.

\bibitem[Tseng et~al.(2021{\natexlab{a}})Tseng, Colburn, Whitehead, Huang,
  Baek, Majumdar, and Heide]{tseng2021neural}
E.~Tseng, S.~Colburn, J.~Whitehead, L.~Huang, S.-H. Baek, A.~Majumdar, and
  F.~Heide.
\newblock Neural nano-optics for high-quality thin lens imaging.
\newblock \emph{Nature communications}, 12\penalty0 (1):\penalty0 1--7,
  2021{\natexlab{a}}.

\bibitem[Tseng et~al.(2021{\natexlab{b}})Tseng, Mosleh, Mannan, St-Arnaud,
  Sharma, Peng, Braun, Nowrouzezahrai, Lalonde, and
  Heide]{tseng2021differentiable}
E.~Tseng, A.~Mosleh, F.~Mannan, K.~St-Arnaud, A.~Sharma, Y.~Peng, A.~Braun,
  D.~Nowrouzezahrai, J.-F. Lalonde, and F.~Heide.
\newblock Differentiable compound optics and processing pipeline optimization
  for end-to-end camera design.
\newblock \emph{ACM Trans. Graph.}, 40\penalty0 (2):\penalty0 1--19, jun
  2021{\natexlab{b}}.
\newblock ISSN 0730-0301.
\newblock \doi{10.1145/3446791}.

\bibitem[Turnhout and Bociort(2009)]{turnhout2009instabilities}
M.~v. Turnhout and F.~Bociort.
\newblock Instabilities and fractal basins of attraction in optical system
  optimization.
\newblock \emph{Optics Express}, 17\penalty0 (1):\penalty0 314--328, Jan. 2009.
\newblock ISSN 1094-4087.
\newblock \doi{10.1364/OE.17.000314}.

\bibitem[{Udacity}(2022)]{udacity2022annotated}
{Udacity}.
\newblock Annotated driving dataset, 2022.
\newblock URL
  \url{https://github.com/udacity/self-driving-car/tree/master/annotations}.

\bibitem[Van Den~Oord et~al.(2017)Van Den~Oord, Vinyals, et~al.]{van2017neural}
A.~Van Den~Oord, O.~Vinyals, et~al.
\newblock Neural discrete representation learning.
\newblock \emph{Advances in neural information processing systems}, 30, 2017.

\bibitem[van Turnhout and Bociort(2009)]{vanturnhout2009chaotic}
M.~van Turnhout and F.~Bociort.
\newblock Chaotic behavior in an algorithm to escape from poor local minima in
  lens design.
\newblock \emph{Optics Express}, 17\penalty0 (8):\penalty0 6436--6450, Apr.
  2009.
\newblock ISSN 1094-4087.
\newblock \doi{10.1364/OE.17.006436}.

\bibitem[Volatier et~al.(2017)Volatier, Mendui{\~n}a-Fern{\'a}ndez, and
  Erhard]{volatier2017generalization}
J.-B. Volatier, {\'A}.~Mendui{\~n}a-Fern{\'a}ndez, and M.~Erhard.
\newblock Generalization of differential ray tracing by automatic
  differentiation of computational graphs.
\newblock \emph{Journal of the Optical Society of America A}, 34\penalty0
  (7):\penalty0 1146, July 2017.
\newblock ISSN 1084-7529, 1520-8532.
\newblock \doi{10.1364/JOSAA.34.001146}.

\bibitem[Wang et~al.(2021)Wang, Chen, and Heidrich]{wang2021lens}
C.~Wang, N.~Chen, and W.~Heidrich.
\newblock Lens design optimization by back-propagation.
\newblock In \emph{International {Optical} {Design} {Conference} 2021}, volume
  12078, pages 312--318. SPIE, Nov. 2021.
\newblock \doi{10.1117/12.2603675}.

\bibitem[Wynne(1959)]{wynne1959lens}
C.~G. Wynne.
\newblock Lens {Designing} by {Electronic} {Digital} {Computer}: {I}.
\newblock \emph{Proceedings of the Physical Society}, 73\penalty0 (5):\penalty0
  777--787, May 1959.
\newblock ISSN 0370-1328.
\newblock \doi{10.1088/0370-1328/73/5/310}.

\bibitem[Yabe(2018)]{yabe2018optimization}
A.~Yabe.
\newblock \emph{Optimization in Lens Design}.
\newblock SPIE Press, 2018.

\bibitem[Yu et~al.(2020)Yu, Chen, Wang, Xian, Chen, Liu, Madhavan, and
  Darrell]{yu2020bdd100k}
F.~Yu, H.~Chen, X.~Wang, W.~Xian, Y.~Chen, F.~Liu, V.~Madhavan, and T.~Darrell.
\newblock Bdd100k: A diverse driving dataset for heterogeneous multitask
  learning.
\newblock In \emph{Proceedings of the IEEE/CVF conference on computer vision
  and pattern recognition}, pages 2636--2645, 2020.

\bibitem[{Zemax}(2019)]{zemax2019user}
{Zemax}.
\newblock Zemax {OpticStudio} 19.8 {User} {Manual}.
\newblock Technical report, Zemax, 2019.

\end{thebibliography}
